\documentclass{article}

\usepackage{arxiv}

\usepackage[utf8]{inputenc} 
\usepackage[T1]{fontenc}    
\usepackage{hyperref}       
\usepackage{url}            
\usepackage{booktabs}       
\usepackage{amsmath,amsfonts,amssymb}
\usepackage{nicefrac}       
\usepackage{microtype}      
\usepackage{graphicx}
\usepackage[numbers,sort&compress]{natbib}
\usepackage{doi}
\usepackage{textcomp}
\usepackage{array,makecell,multirow,tabularx}

\title{OCTR2A: Real-Time RGB-to-BEV Trajectory Projection and Forecasting at Urban Intersections}

\author{
  Emanuele Iacobelli \qquad Simone Teglia \qquad Francesco Pro \qquad Claudia Melis Tonti \qquad Irene Amerini \\[4pt]
  Department of Computer, Control and Management Engineering ``Antonio Ruberti''\\
  Sapienza University of Rome, 00185 Rome, Italy\\
  \texttt{\{iacobelli, teglia, pro, melistonti, amerini\}@diag.uniroma1.it}
}

\renewcommand{\shorttitle}{OCTR2A: Real-Time RGB-to-BEV Trajectory Projection and Forecasting}

\hypersetup{
pdftitle={OCTR2A: Real-Time RGB-to-BEV Trajectory Projection and Forecasting at Urban Intersections},
pdfauthor={Emanuele Iacobelli, Simone Teglia, Francesco Pro, Claudia Melis Tonti, Irene Amerini},
pdfkeywords={Trajectory prediction, bird's-eye-view perception, roadside multi-camera systems, cross-view association, urban intersection safety, synthetic benchmarks},
}

\begin{document}
\maketitle

\begin{abstract}
Trajectory forecasting at urban intersections is often studied assuming that reliable bird's-eye-view (BEV) motion histories are available, which in practice requires costly resources such as LiDAR, aerial views, HD maps, or calibrated infrastructure. We address this upstream perception-to-forecasting problem by reconstructing and predicting agents' BEV trajectories only from paired opposite-corner RGB cameras mounted on existing urban infrastructure at variable heights at inference time. We introduce the Opposite-Corner Trajectory Reconstruction and Anticipation Architecture (OCTR2A), a modular pipeline integrating per-view detection with class refinement, geometric cross-camera matching, transformer-based BEV projection, and egocentric GRU forecasting, trained in cascade and mainly supervised with a soft multi-scale spatial-overlap loss. Since no public benchmark provides this sensing configuration, we introduce R3ST++, a Blender-based synthetic benchmark built from SIND trajectories that provides synchronized opposite-corner views, cross-view correspondences, and BEV ground truth, while also serving as a deployment-support simulator for site-specific fine-tuning. Experiments on these datasets show that, with 12 observed and 12 predicted frames at 10\,Hz, OCTR2A achieves projection~-~forecasting ADE/FDE of 0.144/0.140~-~0.154/0.241~meters in real time on a single consumer GPU. Evaluated on ground-truth BEV histories, the forecasting module improves over the strongest state-of-the-art methods on SIND by 66.7\%/41.7\% ADE/FDE. Transfer experiments on TUMTraf-V2X show that R3ST++ pretraining supports adaptation to real-world roadside domains with different sensing configurations.
\end{abstract}


\section{Introduction}
\label{s:intro}

Recovering and forecasting the trajectories of neighboring agents is a core feature for cooperative intelligent transportation systems and autonomous driving. At signalized urban intersections, this task is particularly challenging because agents enter the same conflict area from multiple directions, react to traffic-light phases and right-of-way rules, and move under heterogeneous kinematic constraints. Additionally, in dense mixed traffic, the difficulty further increases \cite{ref2,ref3,ref4}. Under these conditions, trajectory forecasting cannot be treated as an isolated prediction problem: its reliability also depends on the availability of accurate and temporally coherent ground-plane histories in near real time.

Most recent methods assume that such histories are already available and show that bird's-eye-view (BEV) trajectories, traffic-light states, map topology, and interaction graphs substantially improve forecasting performance in complex intersection scenarios. In many forecasting pipelines, this contextual information is encoded through High Definition (HD) maps, that is, centimeter-accurate digital representations of lane geometry, boundaries, and traffic signals. This line of work leaves open an important deployment question for intersection monitoring: how can reliable histories be obtained when only ground-level RGB streams are available? RGB cameras remain attractive for large-scale deployment because of their low cost, simple installation, and compatibility with existing urban infrastructure. However, a single monocular view remains limited for BEV localization: perspective projection introduces depth ambiguity, wide-angle optics may cause geometric distortion, and static roadside viewpoints are vulnerable to occlusions from large vehicles \cite{ref5}. A complementary opposite-corner dual-view configuration can mitigate these limitations by observing agents under different projective distortions, exploiting elevated mounting points (e.g., traffic-light poles, lighting poles, or monitoring infrastructure) commonly available at intersections. At the same time, this configuration introduces a more complex problem, requiring detection, cross-view association, BEV projection, and forecasting to operate coherently under real-time constraints. 

To address this problem, we propose OCTR2A, an Opposite-Corner Trajectory Reconstruction and Anticipation Architecture that reconstructs and forecasts BEV trajectories from synchronized RGB streams acquired by cameras mounted at opposite corners of the intersection, at varying heights, and directed toward its center. Since no public benchmark currently provides the described sensing configuration, we introduce R3ST++, an enhanced and significantly expanded version of the Real 3D Synthetic Trajectories (R3ST) \cite{teglia2025r3st} dataset. R3ST is a Blender-based synthetic data generator that animates real vehicle trajectories from the Signalized INtersection Dataset (SIND) \cite{ref1} from multiple camera perspectives. 

The main contributions of our work can be summarized as: 
\begin{itemize}
    \item We study an opposite-corner dual-camera sensing configuration for recovering and forecasting BEV trajectories at signalized intersections without LiDAR, aerial views, HD maps, traffic-signal inputs, or manually calibrated homographies at inference time.
    \item We introduce OCTR2A, a real-time modular architecture composed of four independently trainable stages: detection with class refinement, cross-camera matching, BEV projection, and BEV forecasting. These stages are trained in cascade using a soft multi-scale spatial-overlap loss that, unlike the displacement-based and likelihood-based losses adopted in prior trajectory-prediction and BEV methods, directly rewards spatial agreement at multiple scales on the BEV ground plane. Per-term ablations identify this term as the most influential for reconstruction and beneficial for forecasting. 
    \item We provide R3ST++, a synthetic dataset and deployment-support simulator that supplies controlled cross-view correspondences and BEV targets while preserving realistic signalized-intersection trajectory patterns. R3ST++ also helps field deployment by generating site-specific fine-tuning data from a small set of coarse installation parameters, such as opposite-corner camera height and distance from the intersection center.
    \item We separately train and evaluate the forecasting module on SIND with k-fold cross-validation, outperforming the strongest state-of-the-art (SOTA) methods on this dataset in both settings with 12-12 and 12-18 observed and predicted frames. 
\end{itemize}

\section{Related Work}
\label{s:rw}

\subsection{Agent Detection and Cross-Camera Matching}
\label{ss:rw_d_ccm}

In roadside analytics, agents are commonly localized from infrastructure-based monitoring (i.e., fixed roadside sensors) or onboard sensors with general-purpose detectors: one-stage models~\cite{liu2016ssd,lin2017focal}, to which the well-known YOLO~\cite{jocher2026ultralyticsyolo26unifiedrealtime} family belongs, two-stage~\cite{ren2015faster} networks, and Transformer-based~\cite{carion2020detr} architectures. Once frame-level detections are obtained, agents are linked over time through tracking-by-detection algorithms (e.g., SORT~\cite{bewley2016sort}, ByteTrack~\cite{zhang2022bytetrack}, or BoT-SORT~\cite{aharon2022botsort}), increasing robustness to occlusion, non-linear motion, and camera motion. 
 
Cross-camera identity assignment has mainly been investigated in appearance-based Re-Identification and Multi-Target Multi-Camera (MTMC) tracking, for persons \cite{ref14,ref15} and vehicles \cite{veri2016,cityflow2019}. In particular, motorized vehicles are more challenging because instances sharing model and color can be nearly indistinguishable. MTMC systems typically reduce the cross-camera search space by combining appearance descriptors with structural constraints such as camera topology, travel-time models, or spatio-temporal clustering \cite{hsu2021mtmct,szucs2024mcmt}. In contrast to these appearance-centric approaches, our setting exploits synchronized overlapping views, allowing association to be solved with purely geometric cues.

\subsection{Camera-to-BEV Transformation}
\label{ss:rw_ctbt}

The camera-to-ground-plane projection is typically solved through planar homographies or inverse perspective mapping \cite{ref20}, which are efficient and interpretable but highly dependent on accurate site-specific calibration, lens undistortion, and the local planarity of the road surface. Alternatives to overcome these limitations include appearance-based single-view BEV generation \cite{prajapati2024sddunet} and height-to-ground regression for elevated cameras \cite{bevheight2023}. 

In autonomous driving, learned view transformation has been studied extensively for calibrated surround-view rigs, from depth-distributed feature lifting \cite{lss2020} to query-based spatio-temporal Transformers and their refinements \cite{ref21,bevdepth2023}, real-time map-view semantic segmentation \cite{cvt2022}, and camera-LiDAR fusion in a shared BEV space \cite{ref23}. Calibrated roadside camera-LiDAR datasets such as TUMTraf \cite{tumtraf2023} confirm the achievable accuracy but also highlight the sensor, calibration, and annotation costs involved. Taken together, these methods are designed either for ego-vehicle platforms or for roadside setups relying on single viewpoints or camera-LiDAR infrastructure. Hence, none targets the opposite-corner calibration-light RGB configuration addressed in this work.

\subsection{Trajectory Forecasting}
\label{ss:tf}

Interaction-aware forecasting first emerged in the pedestrian literature with social pooling associated with adversarial and graph-based formulations \cite{ref26,ref27,ref28}. In mixed urban traffic, autonomous-driving predictors typically consume agent histories already expressed in ground-plane coordinates, exploiting HD maps or lane graphs through vectorized, graph-based, anchor-based, or attention-based encoders \cite{vectornet2020,lanegcn2020,multipathpp2022,mtr2022,qcnet2023}, improving temporal consistency across successive forecasts \cite{ref38}, scene-level joint prediction \cite{ref34}, or dynamics-aware multimodal decoding \cite{ref30}. Since generating and maintaining HD maps requires time, specialized equipment, and such maps are not always available, a complementary line of research predicts trajectories from observed agent states alone, modeling heterogeneous agents and their interactions through recurrent, graph-convolutional, or attention-based architectures \cite{trafficpredict2019,ref31,ref32,ref33}. A further set of works injects infrastructure-derived knowledge, mainly the traffic-signal state, into generative or graph-based predictors at signalized intersections \cite{ref10,ref9}, while scene-centric cooperative models and Infrastructure-to-Everything formulations combine roadside observations, signal states, HD maps, and maneuver priors \cite{ref19,ref6}. All these methods, however, assume that object-level histories and scene attributes are already available in BEV or metric coordinates. Therefore, their robustness when histories must be reconstructed from raw roadside sensing through detection, tracking, and projection errors remains an open issue.

\subsection{Datasets and Perception-to-Forecasting Frameworks}
\label{ss:rw_d_ptff}

Supervising the proposed pipeline requires synchronized multi-view roadside RGB images, cross-view identity associations, and BEV trajectories, a combination that is difficult to obtain at scale from real footage. Simulators such as CARLA \cite{ref35} and SUMO \cite{ref36}, possibly combined with domain randomization \cite{ref13,ref37}, and synthetic benchmarks for segmentation or traffic-signal control \cite{synthia2016,syntrac2024} provide controllable imagery, but are not designed to replay measured real-world trajectories in synchronized roadside views with cross-view correspondence. Ego-vehicle benchmarks \cite{nuscenes2020,argoverse2} and drone-based trajectory datasets \cite{ref24,ref25} provide accurate annotations but no fixed opposite-corner ground-level views.

Additionally, only a few works address the full path from raw observations to future trajectories. Ego-vehicle pipelines jointly perform detection, tracking, and forecasting from LiDAR \cite{faf2018,pnpnet2020} or camera-only inputs \cite{vip3d2023}, up to unified perception-prediction-planning frameworks \cite{uniad2023}, but rely on calibrated surround-view rigs and dense BEV supervision. Infrastructure-based systems are closer to our deployment setting, yet existing full-stack pipelines rely on calibrated fish-eye arrays and stop at trajectory extraction \cite{zou2022fullstack}, on LiDAR-equipped roadside units with vector maps and signal states \cite{v2xseq2023}, or on non-overlapping camera networks predicting image-plane transitions rather than metric BEV trajectories \cite{styles2020mctf,styles2021trajectorytensors}. 

To the best of our knowledge, no prior work predicts metric BEV trajectories from two stationary opposite-corner RGB cameras with variable mounting heights and no additional sensing, mapping, or calibration input at inference time. Moreover, since no existing benchmark supports such a pipeline either, R3ST++ fills this gap under controlled conditions (see Section~\ref{sss:t_sdvd__r}).

\begin{figure}[!htbp]
\centering
\includegraphics[width=0.9\textwidth]{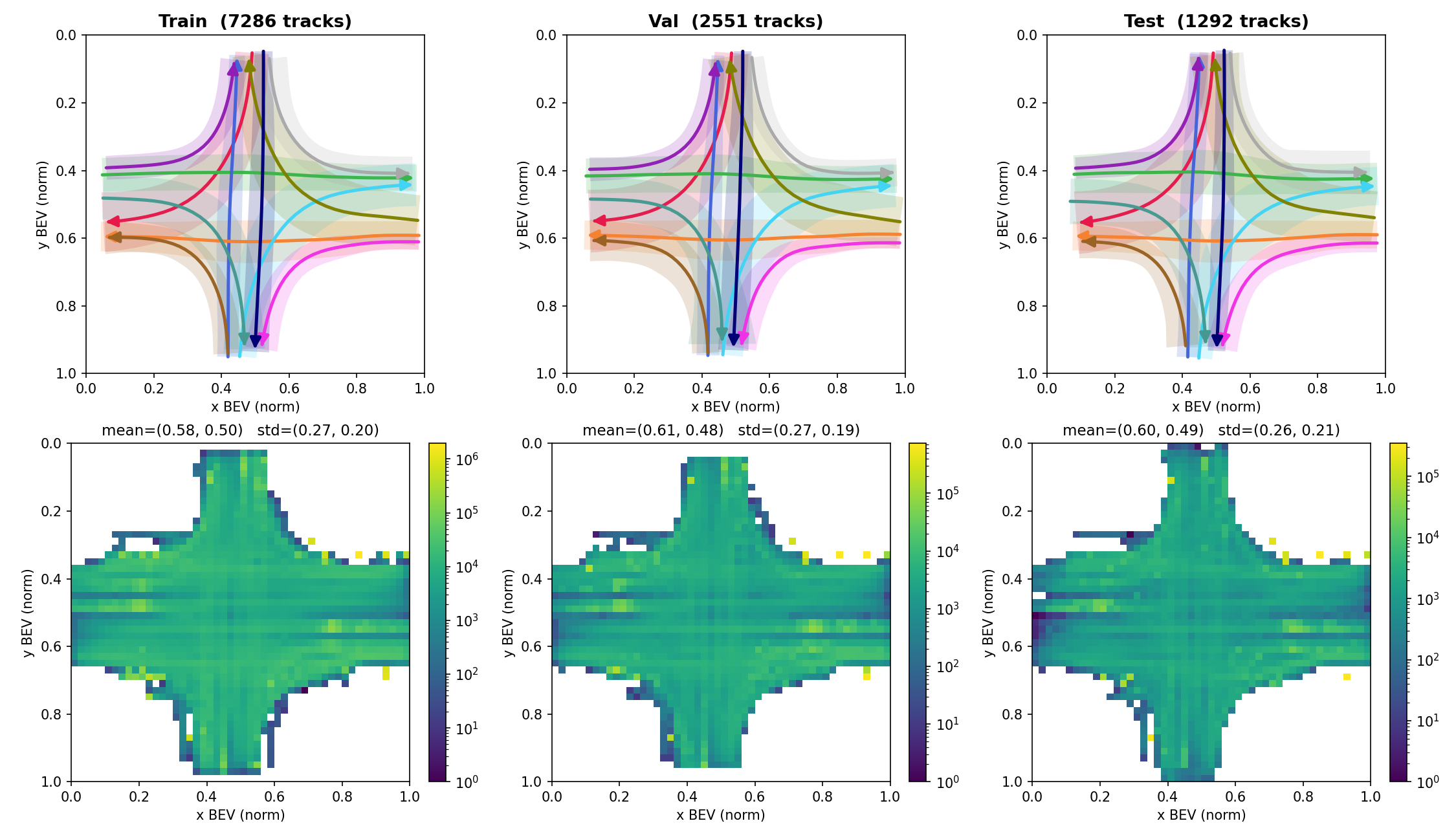}
\caption{BEV track distribution for the fold-0 cross-validation split of R3ST++ (described in Section~\ref{ss:es_rep}), shown in normalized image-plane coordinates for the train, validation, and test subsets. Top: mean trajectories of the 12 maneuvers obtained by K-Means clustering, with $\pm1\sigma$ bands. Bottom: position-density heatmaps computed as 2D histograms in log scale.}
\label{f:sind_trajs}
\end{figure}

\section{Datasets and Evaluation Domains}
\label{s:d_ed}

Four datasets are employed to train and evaluate OCTR2A. VisDrone~\cite{zhu2021detection} supports image-level training of the detection and class-refinement modules. SIND provides the real trajectory source for the motion-centered components. R3ST++ renders trajectories in the target dual-camera setting, providing synchronized cross-view observations, correspondence labels, and BEV supervision for the matching, projection, and forecasting stages. Finally, TUMTraf serves as an external real-roadside benchmark with intersection images acquired under a different sensing geometry.

\subsection{Detection and Classification Dataset}
\label{ss:d_ed_d_cd}

VisDrone is a large-scale drone-based visual benchmark containing images captured in urban and suburban scenarios across 14 Chinese cities under diverse weather, lighting, viewpoint, altitude, and object-density conditions. Its strong scale variation, dense traffic, and frequent occlusions make it suitable for pretraining our detection stage before applying it to synthetic or real-world camera intersection settings. This dataset provides more than 2.5 million annotated bounding boxes for 10 common object classes: pedestrian, person, car, van, bus, truck, motorcycle, bicycle, awning-tricycle, and tricycle. For this research, we selected the Object Detection in Images (VisDrone-DET) challenge dataset, and for consistency with the classes available in the trajectory datasets (Section~\ref{ss:d_ed_t_sdvd}), we merged some labels before training. In particular, the person category is mapped to pedestrian, the awning-tricycle is mapped to tricycle, and the van is mapped to car.

\subsection{Trajectory and Synthetic Dual-View Datasets}
\label{ss:d_ed_t_sdvd}

\subsubsection{SIND}
\label{sss:t_sdvd__s}

It is a drone-based trajectory dataset collected at four two-phase signalized intersections in China, providing synchronized traffic-light states, agent trajectories, and HD maps in lanelet2 \cite{poggenhans2018lanelet2} format. The main recordings cover approximately 7 hours of traffic over 23 sessions in the city of Tianjin, yielding trajectories at 10~Hz for 13{,}248 participants across 7 classes (pedestrian, car, bus, truck, motorcycle, bicycle, and tricycle) with position, velocity, acceleration, size, and heading (pedestrians excluded). In this work, we used SIND as both the real trajectory source for R3ST++ and the benchmark for evaluating future BEV trajectories predicted from ground-truth BEV histories. Fig.~\ref{f:sind_trajs} illustrates the trajectory balance within the dataset, showing consistent maneuver patterns, spatial densities, and summary statistics.

\begin{figure}[!htbp]
\centering
\begin{tabular}{@{}c@{\hspace{1pt}}c@{\hspace{1pt}}c@{}}
\includegraphics[width=0.33\columnwidth]{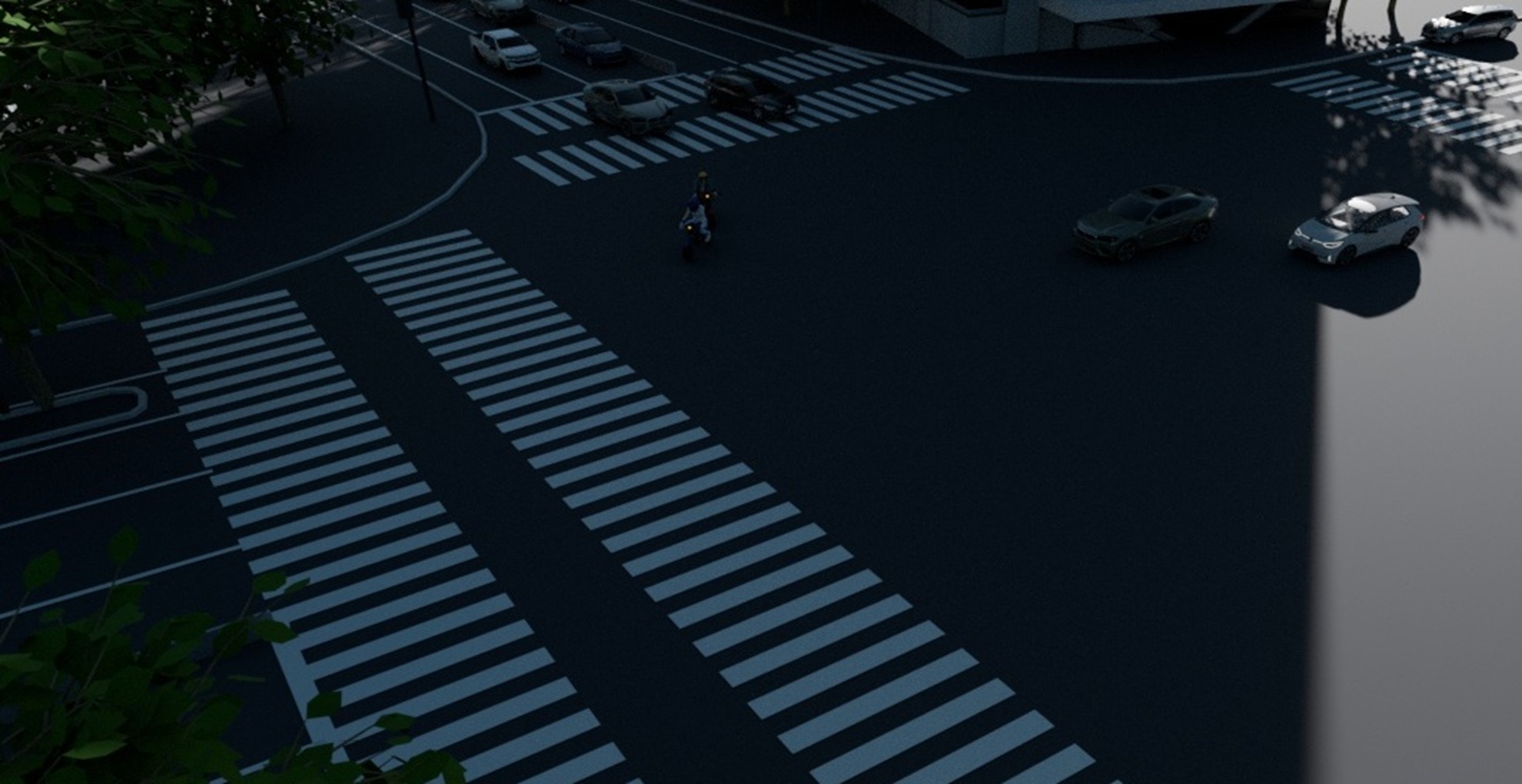} &
\includegraphics[width=0.33\columnwidth]{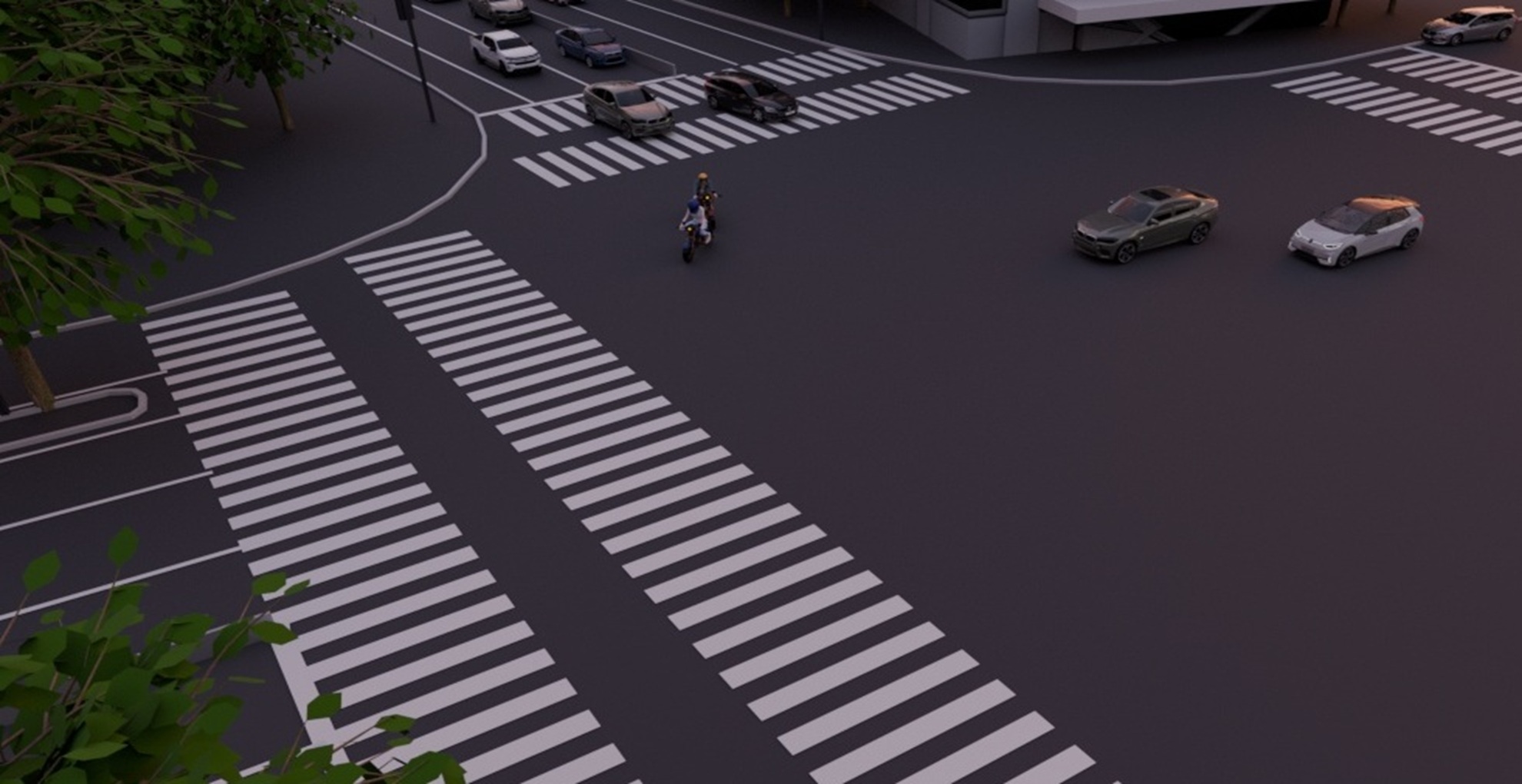} &
\includegraphics[width=0.33\columnwidth]{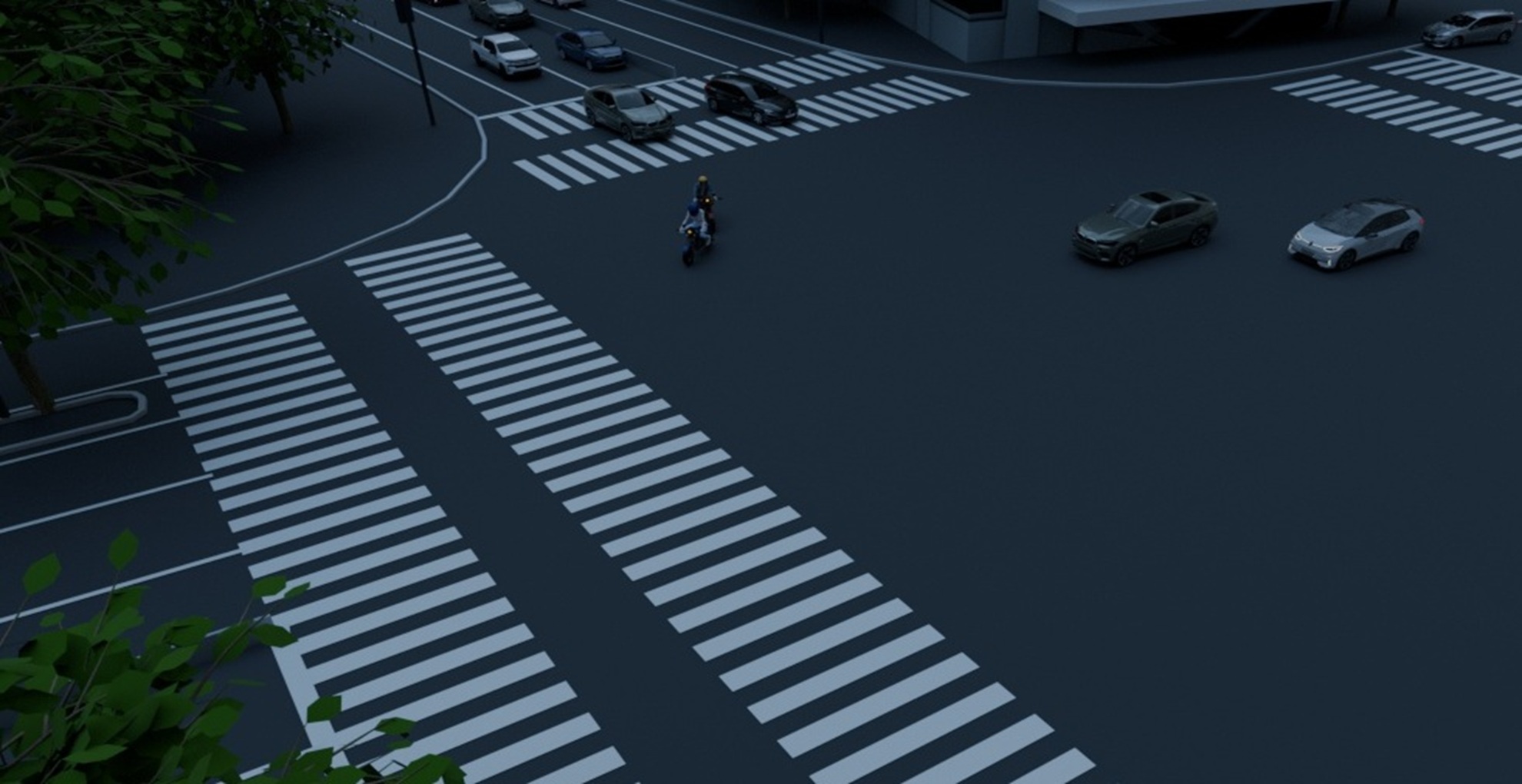} \\
[-0.2em] a) & b) & c) \\ [3pt]
\includegraphics[width=0.33\columnwidth]{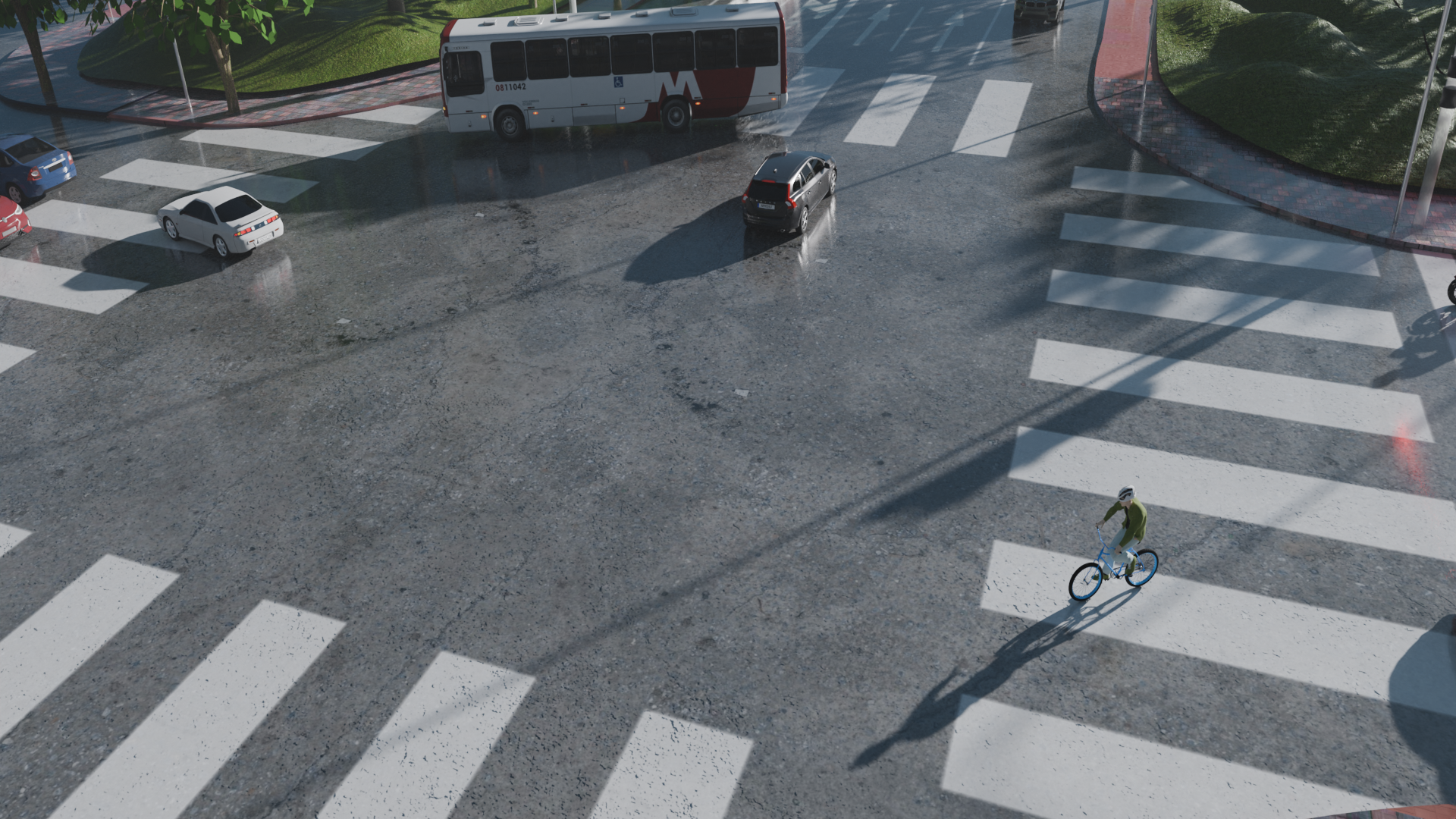} &
\includegraphics[width=0.33\columnwidth]{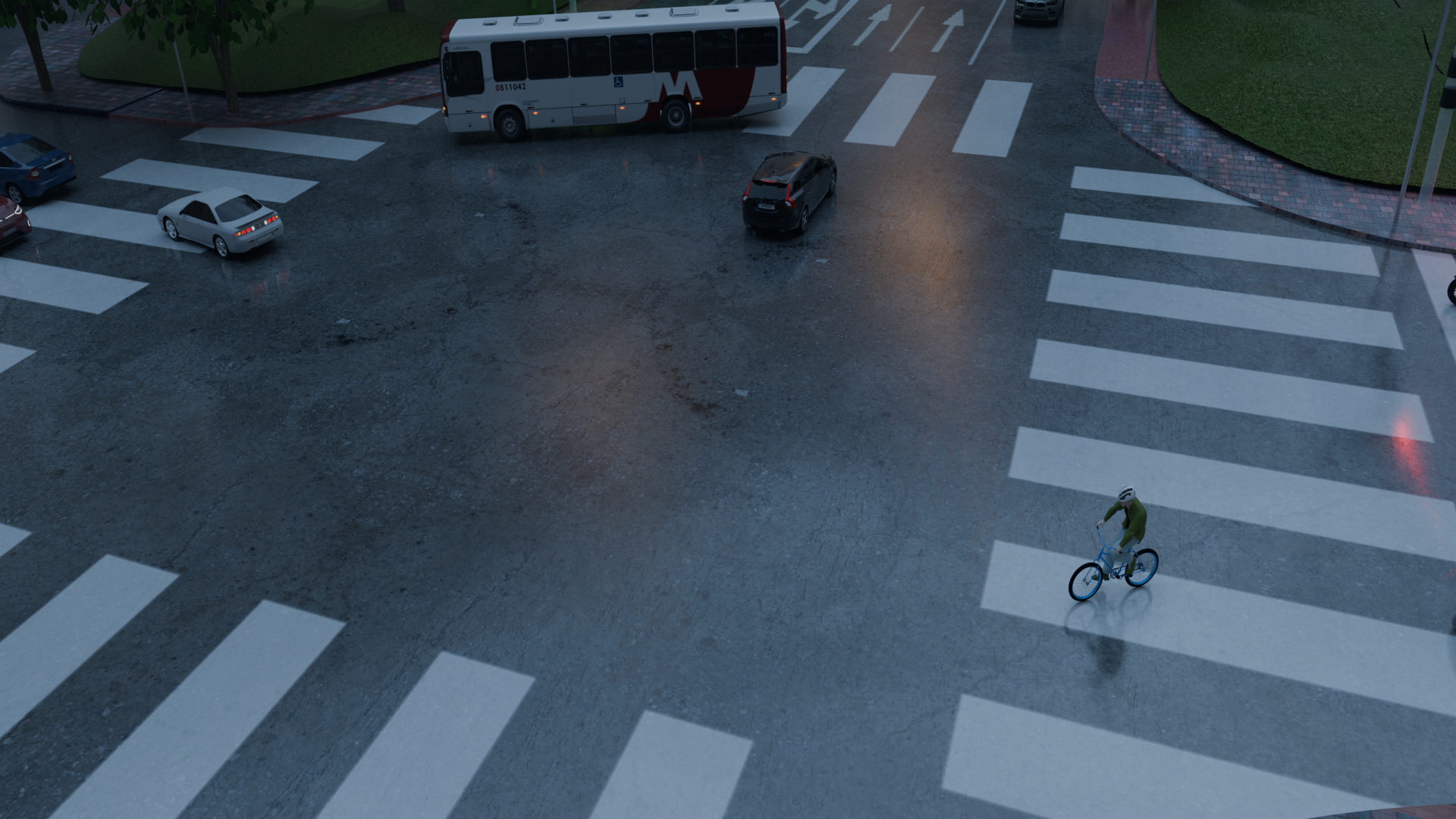} &
\includegraphics[width=0.33\columnwidth]{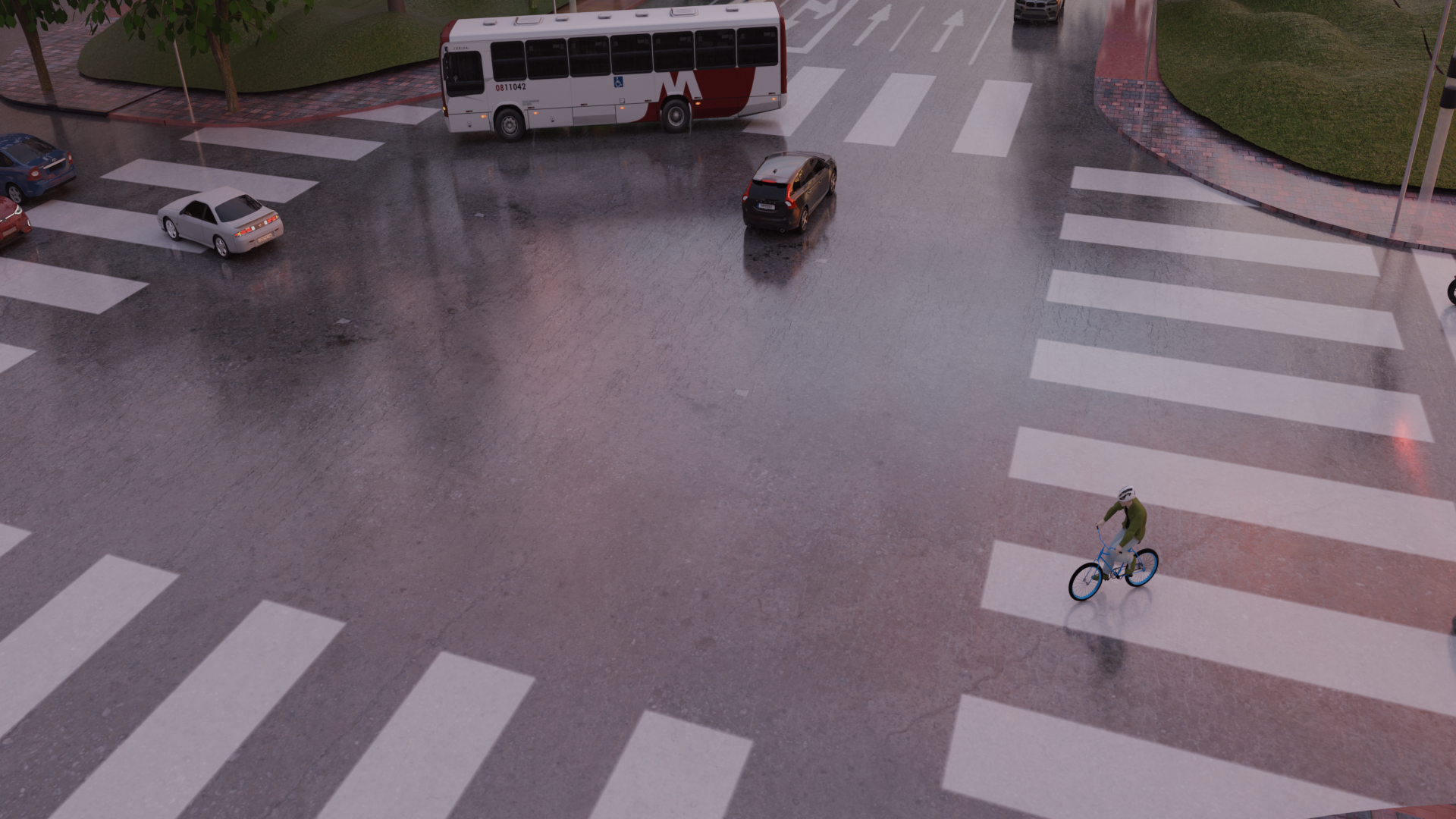} \\
[-0.2em] d) & e) & f) \\
\end{tabular}
\caption{R3ST++ rendering variability across illumination and weather conditions. The top row shows a dry intersection under sunny afternoon (a), sunset (b), and cloudy midday (c) lighting. The bottom row shows a second intersection after rainfall, under the same lighting conditions.}
\label{fig:R3ST_environment_lights}
\end{figure}

\begin{figure}[!htbp]
\centering
\begin{tabular}{@{}c@{\hspace{1pt}}c@{}}
\includegraphics[width=0.4\textwidth]{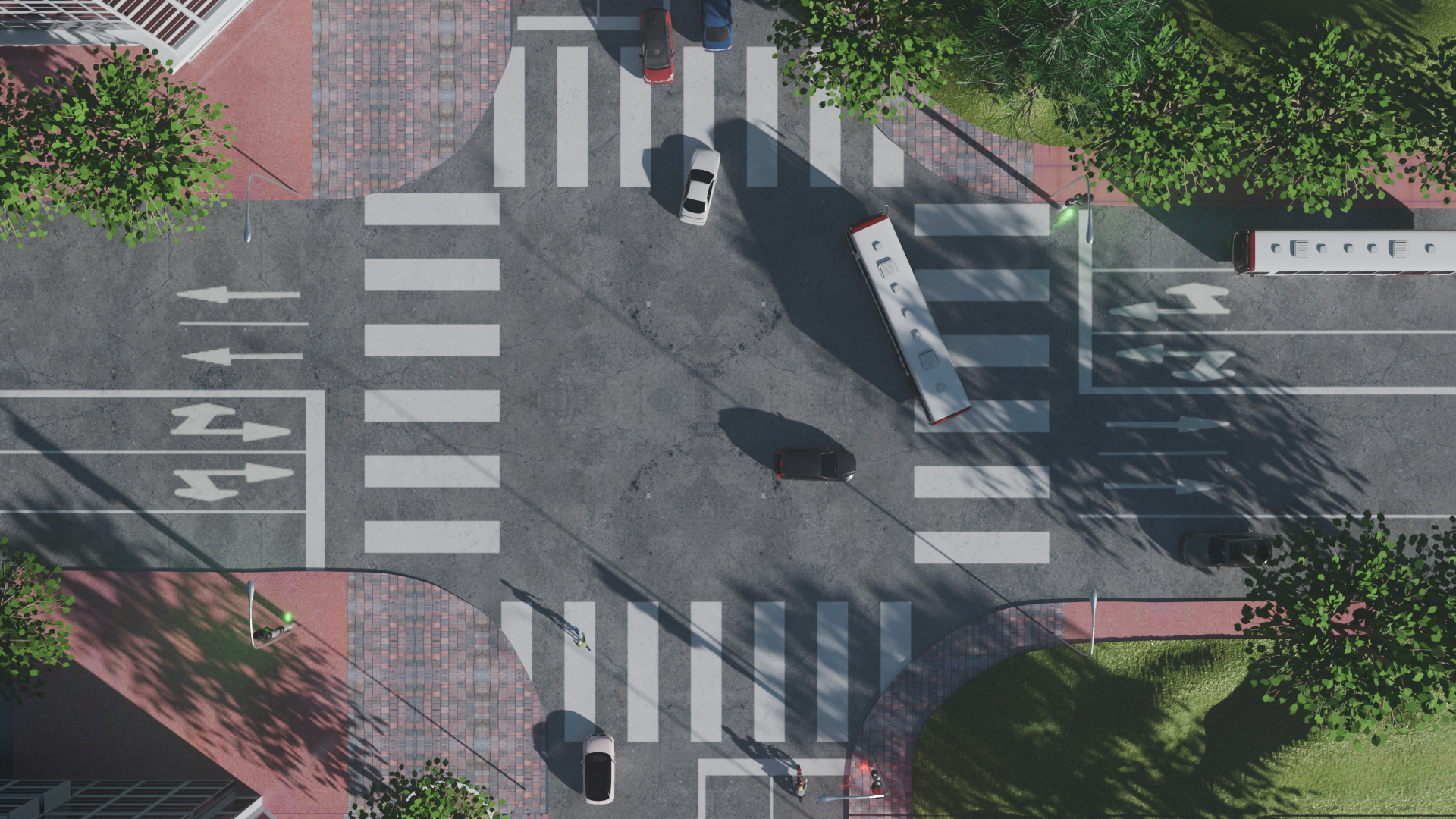} &
\includegraphics[width=0.4\textwidth]{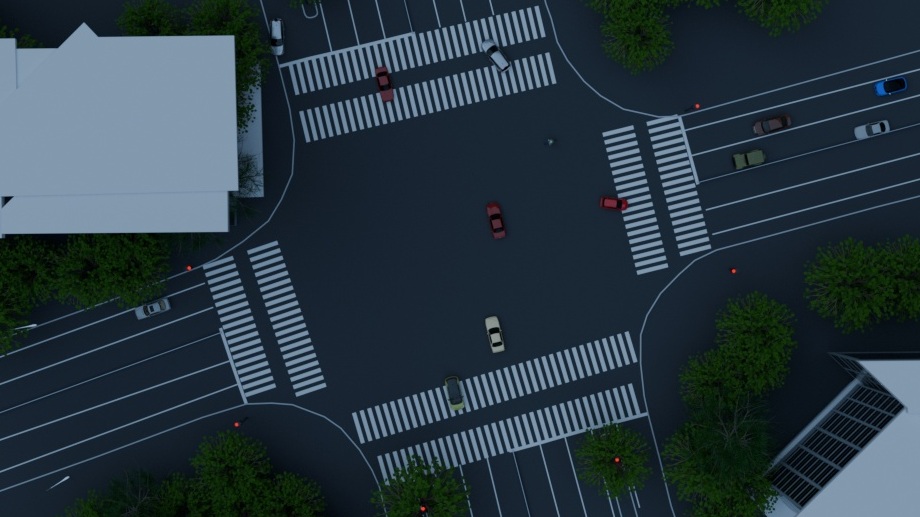} \\
[-0.2em] a) & b) \\
\end{tabular}
\caption{Examples of intersection topologies in R3ST++: the city of Tianjin (a), and Xi'an (b).}
\label{fig:BEV_R3ST_topologies}
\end{figure}

\subsubsection{R3ST++}
\label{sss:t_sdvd__r}

It builds on R3ST, a Blender-based simulation framework that animates real traffic recordings inside configurable 3D intersection scenes. It provides the controlled sensing conditions needed to supervise cross-camera matching, BEV reconstruction, and forecasting, while avoiding the artificial driving behaviors often produced by the datasets discussed in Section~\ref{ss:rw_d_ptff}. Additionally, it supplies dense per-frame annotations that extend the dataset for auxiliary tasks such as instance segmentation and monocular depth estimation. Compared to its predecessor, R3ST++ increases the agent heterogeneity to 70 unique 3D actors distributed across seven main classes: pedestrian, car, bus, truck, motorcycle, bicycle, and tricycle. These 3D assets are collected from multiple sources, including Blendkit \cite{blenderkit} and Sketchfab \cite{sketchfab}, while some others are manually designed (e.g., persons riding motorcycles). Environmental variability is also enhanced through three lighting configurations (midday, afternoon, and sunset) plus optional fog or rain (see Fig.~\ref{fig:R3ST_environment_lights}). R3ST++ further extends the structural diversity of the benchmark by supporting distinct intersection topologies, including different road geometries, traffic flows, and right-of-way configurations (see Fig.~\ref{fig:BEV_R3ST_topologies}). In addition, it supports multi-view camera synthesis with adjustable parameters, enabling the simulation of different sensor placement strategies.

In this work, the 3D model of the Tianjin intersection of SIND is selected. At each timestep, class-specific 3D agent models are instantiated and animated according to the ground-truth trajectories. Four virtual cameras are placed at the corners of the intersection, at realistic mounting heights of approximately 10 meters (representative of traffic-monitoring pole installations), and oriented to provide complementary coverage of the intersection center. An additional overhead camera is placed above the scene to render BEV supervision over the full spatial extent of the trajectories. For each timestep, synchronized frames are rendered from two opposite-corner cameras and the overhead camera. The 2D agent annotations are obtained by projecting the 3D bounding volumes through the corresponding virtual projection matrices and are normalized to $[0,1]$ in each image plane.

\begin{figure}[!htbp]
\centering
\includegraphics[width=0.55\textwidth]{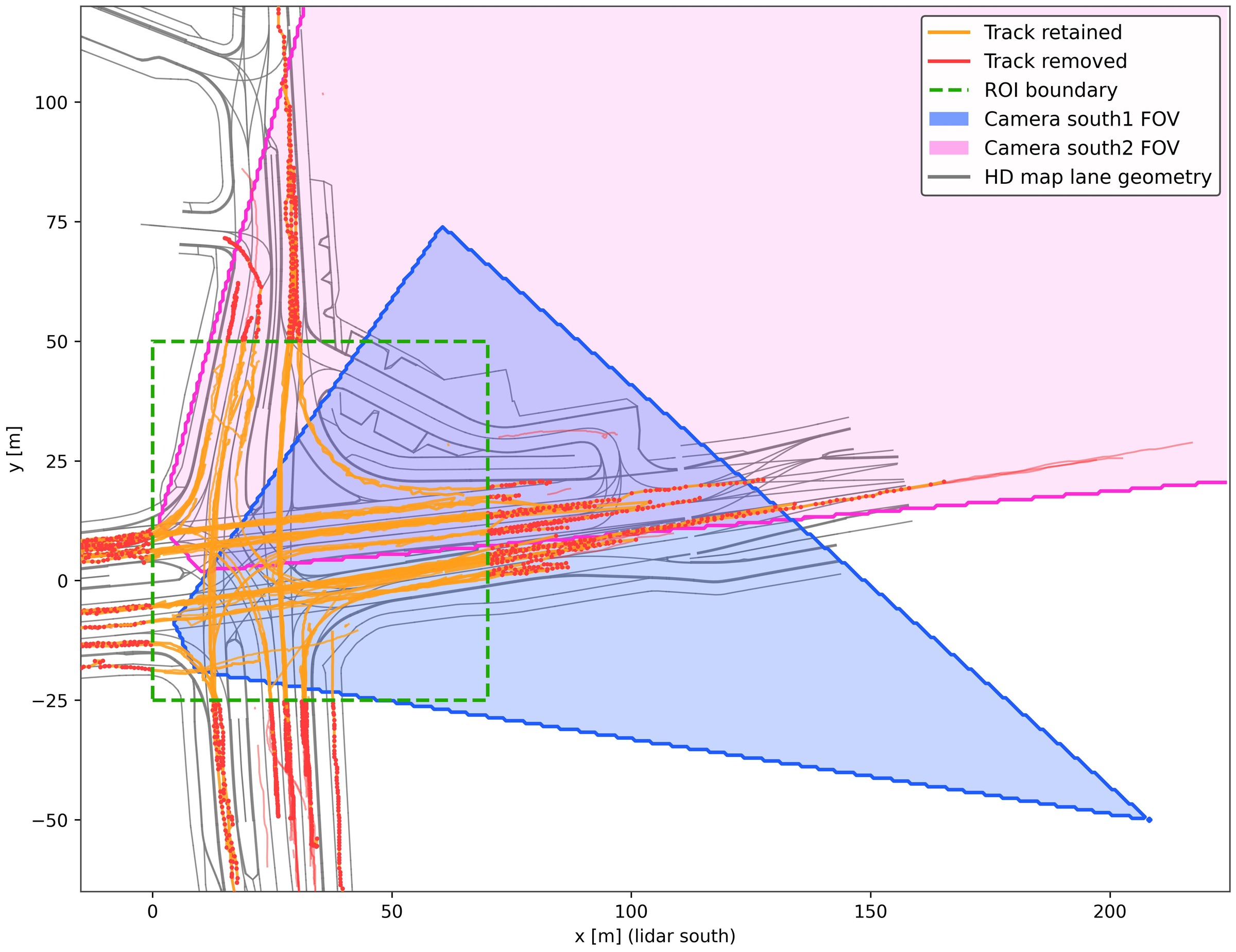}
\caption{TUMTraf-V2X BEV ROI (defined in Section \ref{sss:tumtraf_eval_protocol}) used for the external roadside-domain evaluation, showing retained trajectories within the green box, discarded trajectories, south1/south2 camera FOV, and HD-map lane geometry in metric coordinates.}
\label{fig:roi_tumtraf}
\end{figure}

\subsection{External Real-Roadside Evaluation Domain}
\label{sec:dataset_tumtraf}

TUMTraf is a family of roadside multi-sensor datasets collected in Munich, Germany, providing infrastructure-mounted and onboard camera and LiDAR data with calibration information, 3D bounding boxes, and track identifiers, covering highway, intersection, event-based, and cooperative perception scenarios. For this paper, we use TUMTraf-V2X~\cite{zimmer2024tumtraf}, a dataset designed for cooperative 3D object detection and tracking. Although its scenarios are deliberately challenging for single-camera perception, with frequent occlusions that make LiDAR information important, it remains the dataset whose roadside sensing configuration most closely resembles the one targeted by our framework. Specifically, we select the south1 and south2 RGB cameras, which observe a partially overlapping region of the intersection (see Fig.~\ref{fig:roi_tumtraf}), and derive BEV trajectories from the LiDAR-based 3D annotations. The dataset provides $9$ agent classes: car, truck, trailer, van, motorcycle, bus, pedestrian, bicycle, and emergency. For consistency with the previously described datasets, van and emergency are mapped to car, while trailer is merged with the associated truck into a single box.
The adaptations we made to apply our architecture are described in Section~\ref{sss:tumtraf_eval_protocol}.

\begin{figure*}[!htbp]
\centering
\includegraphics[width=\textwidth]{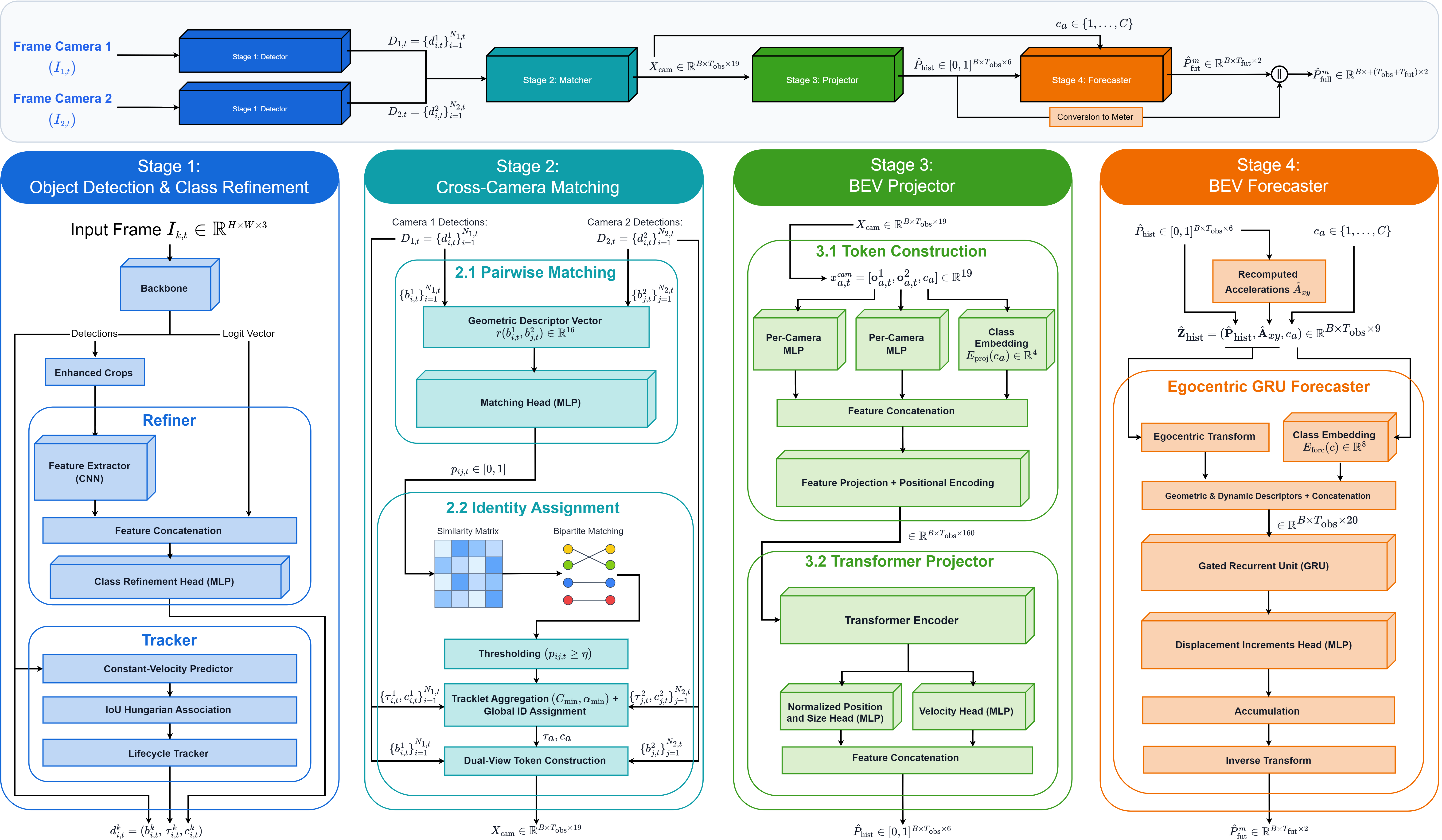}
\caption{Overview of OCTR2A: two opposite-corner RGB streams are processed by per-view detection with class refinement (Stage~1), geometric cross-camera matching (Stage~2), Transformer-based BEV projection (Stage~3), and egocentric GRU forecasting (Stage~4), yielding metric BEV histories and future trajectories.}
\label{fig:overview_architecture}
\end{figure*}

\section{Methodology}
\label{Methodology}

This section describes OCTR2A from a technical perspective, focusing on the notation, module implementation, and training objectives that define the complete pipeline. An overview of our end-to-end framework is shown in Fig.~\ref{fig:overview_architecture}.

\subsection{Problem Formulation}
\label{subsec:problem_definition}

Let $\mathbf{I}_{k,t} \in \mathbb{R}^{H \times W \times 3}$ ($k \in \{1,2\}$) denote the RGB frame from camera $k$ at timestep $t$. For each frame, the detection stage returns a set of camera-local observations $\mathcal{D}_{k,t} = \{d^{k}_{i,t}\}_{i=1}^{N_{k,t}}$, where $N_{k,t}$ is the number of detected agents in camera $k$. Each detection $d^{k}_{i,t} = (\mathbf{b}^{k}_{i,t}, c^{k}_{i,t}, \tau^{k}_{i,t})$ consists of a bounding box $\mathbf{b}^{k}_{i,t} = (x^{k}_{i,t}, y^{k}_{i,t}, w^{k}_{i,t}, h^{k}_{i,t})$ (defined by its center coordinates, width, and height), a scalar class label $c^{k}_{i,t}$, and a single-view track identifier $\tau^{k}_{i,t}$.

After cross-camera association, detections of the same physical agent are grouped into an identity $a$. For each observed timestep, the camera-level observation of agent $a$ in view $k$ is
\begin{equation}
\mathbf{o}^{k}_{a,t} = \left[\mathbf{b}^{k}_{a,t}, \dot{x}^{k}_{a,t}, \dot{y}^{k}_{a,t}, \ddot{x}^{k}_{a,t}, \ddot{y}^{k}_{a,t}, \nu^{k}_{a,t}\right]^{\top} \in \mathbb{R}^{9},
\end{equation}
where velocity/acceleration are computed in the image plane coordinates and $\nu^{k}_{a,t}\in\{0,1\}$ flags the agent's visibility in camera $k$. Concatenating both views with the shared class label $c_a$ gives the dual-camera token $\mathbf{x}^{\mathrm{cam}}_{a,t} = [\mathbf{o}^{1}_{a,t}, \mathbf{o}^{2}_{a,t}, c_a] \in \mathbb{R}^{19}$. Stacking over the observed window $T_{\mathrm{obs}}$ yields the \textit{projector} input $\mathbf{X}_{\mathrm{cam}} \in \mathbb{R}^{B \times T_{\mathrm{obs}} \times 19}$ for a batch $B$ of agents. 

The BEV \textit{projector} $f_\theta$ maps the dual-camera history to BEV history states:
\begin{equation}
\hat P_{\mathrm{hist}}=f_\theta(X_{\mathrm{cam}})\in\big([0,1]^{4}\times\mathbb{R}^{2}\big)^{B\times T_{\mathrm{obs}}},
\label{eq:projector_problem}
\end{equation}
with each predicted state:
\begin{equation}
\hat{\mathbf{p}}^{\mathrm{BEV}}_{a,t}
=
\left(
\hat{x}^{\mathrm{BEV}}_{a,t},
\hat{y}^{\mathrm{BEV}}_{a,t},
\hat{w}^{\mathrm{BEV}}_{a,t},
\hat{h}^{\mathrm{BEV}}_{a,t},
\hat{v_x}^{\mathrm{BEV}}_{a,t},
\hat{v_y}^{\mathrm{BEV}}_{a,t}
\right)
\end{equation} 
where position and size are normalized according to the dimensions of the BEV camera's image plane, while the velocities are signed.

The BEV \textit{forecaster} $g_{\phi}$ receives as input the BEV history, the recomputed accelerations, and the scalar class as $\hat{\mathbf{Z}}_{\mathrm{hist}}=(\hat{\mathbf{P}}_{\mathrm{hist}}, \hat{\mathbf{A}}_{xy}, c_a) \in \mathbb{R}^{B \times T_{\mathrm{obs}} \times 9}$, and predicts future BEV positions in meters over the horizon $T_{\mathrm{fut}}$:
\begin{equation}
\hat{\mathbf{P}}^{m}_{\mathrm{fut}} = g_{\phi}(\hat{\mathbf{Z}}_{\mathrm{hist}}) \in \mathbb{R}^{B \times T_{\mathrm{fut}} \times 2}.
\label{eq:forecaster_problem}
\end{equation}

The final OCTR2A output concatenates the projected observed history (converted to meters) with the predicted future trajectory:
\begin{equation}
\hat{\mathbf{P}}^{m}_{\mathrm{full}} = \left[\hat{\mathbf{P}}^{m}_{\mathrm{hist}}, \hat{\mathbf{P}}^{m}_{\mathrm{fut}}\right] \in \mathbb{R}^{B \times (T_{\mathrm{obs}}+T_{\mathrm{fut}}) \times 2}.
\label{eq:full_bev_trajectory}
\end{equation}

\subsection{Pipeline Overview}
\label{ss:pipeline_overview}

\subsubsection{Object Detection and Classification}

The first stage, the \textit{detector}, processes the two RGB camera streams independently and aims to detect traffic agents while assigning them a reliable semantic class. Each image is analyzed first by a \textit{backbone} network (see Section~\ref{sss_r3st_train_imp_det}), which localizes the objects of interest and produces, for every detected bounding box, an initial class prediction and the associated class-logit vector.

Since small, distant, or partially occluded road users may be difficult to classify accurately from the \textit{backbone} output alone, each detection is further processed by a CNN-based \textit{refiner} network. For every detected object, the original bounding box is upscaled by 20~\% to include additional local context around the agent. The resulting crop is resized to $64\times64$ pixels and passed to the \textit{refiner}, which consists of three convolutional blocks (24/48/96 channel widths), each with a $3\times3$ kernel and followed by batch normalization, SiLU activation, and $2\times2$ max pooling. A final global average pooling layer then produces a 96-dimensional visual descriptor that is concatenated with the $C$-dimensional \textit{backbone} class-logit vector. The fused representation is finally processed by a small Multi-Layer Perceptron (MLP) with hidden dimensions 128 and 64, dropout rate 0.2, and a final $C$-way classification head. It is important to note that the \textit{refiner}
($\approx$75k parameters with $C=7$) only updates the semantic labels, while the bounding-box geometry remains unchanged from the \textit{backbone} output.

Finally, each view is processed by a SORT-like \textit{tracker} that links detections over time and assigns the single-view identifier $\tau^{k}_{i,t}$. For every active track, a constant-velocity model predicts the current bounding-box center, and the predicted boxes are matched to the current detections through a one-to-one Hungarian assignment on the IoU matrix. Low-IoU pairs are discarded, unmatched detections initialize new tracks, and unmatched tracks are retained for a fixed number of frames (see Section~\ref{sss_r3st_train_imp_det}). The association is purely geometric, ignoring the semantic class $c^{k}_{i,t}$, and adds no learnable parameters.

\subsubsection{Cross-Camera Matching}

The second stage, the \textit{matcher}, associates detections of the same physical agent across the two synchronized camera views. Since the video streams observe a partially overlapping region of the intersection at the same timestep, an agent at a given BEV position is expected to generate a geometrically coherent pair of image-plane detections. Hence, the \textit{matcher} compares candidate bounding boxes across views based on their position, displacement, and shape. 

In detail, at timestep $t$, all candidate detection pairs $\left(d^{1}_{i,t}, d^{2}_{j,t}\right) \in \mathcal{D}_{1,t} \times \mathcal{D}_{2,t}$ are evaluated through a 16-dimensional geometric descriptor $\mathbf{r}(\mathbf{b}^{1}_{i,t}, \mathbf{b}^{2}_{j,t})$ which includes bounding-box coordinates and sizes from both views, signed and absolute center displacements, Euclidean inter-box distance, size-ratio and area-ratio features. This descriptor is passed to the \textit{matcher} ($m_{\psi}$) network, which estimates the probability that the two detections correspond to the same physical agent:
\begin{equation}
p_{ij,t}
=
m_{\psi}
\left(
\mathbf{r}\left(\mathbf{b}^{1}_{i,t},\mathbf{b}^{2}_{j,t}\right)
\right)
\in[0,1].
\end{equation}
The \textit{matcher} is a compact MLP with hidden dimensions $128$, $64$, and $32$. Each hidden layer is followed by batch normalization, ReLU activation, and dropout with probability $0.3$ ($\approx$13k parameters). The output layer produces the pairwise matching probability $p_{ij,t}$ through a sigmoid activation.

Cross-camera association is formulated as a partial one-to-one assignment problem, since some agents may be visible in only one camera. Let $z_{ij,t}\in\{0,1\}$ indicate whether detections $i$ and $j$ are matched. The frame-level association follows a solve-then-threshold procedure. First, a maximum-probability rectangular assignment is calculated:
\begin{equation}
\mathbf{Z}^{\star}_t
=
\operatorname*{arg\,max}_{\substack{
\mathbf{Z}_t,\; z_{ij,t}\in\{0,1\}\\}}
\sum_{i=1}^{N_{1,t}}
\sum_{j=1}^{N_{2,t}}
p_{ij,t}z_{ij,t},
\end{equation}
and subject to
{
\begin{equation}
\sum_j z_{ij,t}\leq 1,\qquad
\sum_i z_{ij,t}\leq 1,\qquad
\sum_{i,j}z_{ij,t}=\min(N_{1,t},N_{2,t}).
\end{equation}
}
This problem is solved using the rectangular Hungarian algorithm with cost \(1-p_{ij,t}\). Then, we retain only assignments whose
probability exceeds the confidence threshold:
\begin{equation}
\mathcal{M}_t
=
\left\{
(i,j):z^\star_{ij,t}=1,\;
p_{ij,t}\geq\eta
\right\}.
\end{equation}

Frame-level matches are then aggregated over time to obtain persistent cross-camera identities. For a candidate track pair $(\tau^1,\tau^2)$, let $C(\tau^1,\tau^2)$ be the number of co-visible frames and $V(\tau^1,\tau^2)$ the number of frames in which the two identifiers are linked. A persistent cross-camera identity is accepted only if
\begin{equation}
C(\tau^1,\tau^2) \geq C_{\min},
\qquad
\frac{V(\tau^1,\tau^2)}{C(\tau^1,\tau^2)} \geq \alpha_{\min}.
\end{equation}
Eligible track pairs are ultimately resolved via a global bipartite assignment, yielding shared cross-camera identifiers while preserving unmatched single-view tracks whenever no reliable correspondence is available. A single semantic class $(c_a)$ is then assigned to each resulting identity by majority vote over the detector classes $(c^{k}_{i,t})$ associated with its detections, pooled across both views and over the entire observation window. The vote is unweighted because the detections do not include confidence scores, and it reduces the effect of isolated misclassifications occurring in a single frame or view.

\subsubsection{BEV \textit{Projector}}

The third stage, the \textit{projector}, reconstructs the observed ground-plane history from the matched dual-camera tokens. Rather than estimating each BEV point independently, the \textit{projector} combines instantaneous cross-view geometry with temporal evidence over the observed window, predicting for each timestep the normalized BEV coordinates, velocities, and bounding-box size. This allows the model to exploit how image-plane position, scale, and motion evolve jointly in the two synchronized views. Its core is a Transformer encoder operating over the temporal window. Per-camera features are first encoded with a hidden width of $128$, while the scalar integer class label $c_a$ is mapped through a learned embedding $E_{\text{proj}}(c_a)\in\mathbb{R}^{4}$ and concatenated with the resulting per-view features. Each timestep representation is then projected to a $160$-dimensional vector with added sinusoidal positional encodings and processed by a 3-layer Transformer encoder with 4 attention heads, feed-forward width 330, pre-layer normalization, GELU activations, and dropout 0.25. Two separate MLP heads process the temporal descriptors: the first estimates the normalized BEV position and size with a sigmoid activation function constraining the outputs to the normalized BEV domain. The second head regresses the signed BEV velocity components without an activation function. Overall, this stage contains approximately 735k trainable parameters.

\subsubsection{BEV \textit{Forecaster}}

The fourth stage, the \textit{forecaster}, predicts the future motion of each agent from its BEV history: built from ground-truth trajectories during standalone training and from \textit{projector} outputs during cascaded training. To reduce dependence on absolute scene orientation, each input history is expressed in an egocentric frame centered on the last observed position and rotated so that the last valid observed displacement points along the positive $x$-axis. When its magnitude falls below a threshold $\epsilon$ (i.e., for stopped or near-stationary agents), no rotation is applied. The \textit{forecaster} predicts in this local frame the future motion, and the inverse transform maps the prediction back to BEV coordinates. The input sequence is augmented with geometric and dynamic descriptors (aspect ratio, bounding-box area, diagonal length, and speed magnitude) and a separate learned agent-type embedding $E_{\text{forc}}(c)\in\mathbb{R}^{8}$, yielding a $20$-dimensional per-timestep representation. These features are projected to a $128$-dimensional latent space and processed by a two-layer Gated Recurrent Unit (GRU) with hidden size $128$ and dropout $0.1$. Instead of regressing absolute positions, an MLP head estimates future displacement increments, accumulated over the horizon to obtain the trajectory in world meters, which encourages temporally coherent predictions. Overall, the \textit{forecaster} contains approximately 273k trainable parameters.

\subsection{Training Objectives}
\label{subsec:training_objectives}

\subsubsection{Object Detection and Classification}

The \textit{backbone} is optimized with its standard detection objective, including localization, objectness, and classification terms (see Section~\ref{sss_r3st_train_imp_det}). The \textit{refiner} is supervised with class-weighted cross-entropy with label smoothing:
\begin{equation}
\mathcal{L}_{\mathrm{cls}}
=
-
\sum_{c=1}^{C}
w_c\, \tilde{y}_c \log \hat{y}_c,
\end{equation}
where $w_c$ is the weight of class $c$, compensating for class imbalance, and $\tilde{y}_c$ denotes the smoothed target distribution.

\subsubsection{Cross-Camera Matching}

Since positive cross-camera matches, which represent detections of the same physical agent in the two synchronized views, are much rarer than negative candidate pairs, the matching module is trained with weighted binary cross-entropy:
\begin{equation}
\mathcal{L}_{\mathrm{match}}
=
-
w_{\mathrm{pos}}y\log p
-
w_{\mathrm{neg}}(1-y)\log(1-p),
\end{equation}
where $y\in\{0,1\}$ is the matching label, $p$ is the predicted matching probability, and $w_{pos}$, $w_{neg}$ are the weights associated with positive and negative matches.

\subsubsection{BEV Trajectory Objective}

The \textit{projector} and \textit{forecaster} modules are supervised with the same trajectory objective:

{
\begin{equation}
\mathcal{L}_{\mathrm{traj}}
=
w_{\mathrm{a}}\mathcal{L}_{\mathrm{ADE}}
+
w_{\mathrm{v}}\mathcal{L}_{\mathrm{VEL}}
+
w_{\mathrm{f}}\mathcal{L}_{\mathrm{FDE}}
+
w_{\mathrm{o}}\mathcal{L}_{\mathrm{overlap}}.
\end{equation}
}

The Average Displacement Error (ADE) and the Final Displacement Error (FDE) components preserve direct comparability with standard trajectory-prediction metrics. The average-displacement term measures pointwise localization error over the valid timesteps:
\begin{equation}
\mathcal{L}_{\mathrm{ADE}}
=
\frac{1}{\sum_t m_t}
\sum_t
m_t
\left\|
\hat{\mathbf{p}}_t-\mathbf{p}_t
\right\|_2,
\end{equation}
where $m_t$ is the validity mask. The final-displacement term emphasizes the last valid timestep of the supervised horizon:
\begin{equation}
\mathcal{L}_{\mathrm{FDE}}
=
\left\|
\hat{\mathbf{p}}_{T^{\star}}
-
\mathbf{p}_{T^{\star}}
\right\|_2,
\end{equation}
where $T^{\star}$ is the last valid timestep for each trajectory. The velocity-consistency term compares finite-difference motion between consecutive BEV positions:

{
\begin{equation}
\begin{aligned}
\mathcal{L}_{\mathrm{VEL}}
&=
\frac{1}{\sum_t m_t m_{t-1}}
\sum_t
m_t m_{t-1} \\
&\quad
\left\|
(\frac{\hat{\mathbf{p}}_t-\hat{\mathbf{p}}_{t-1}}{\Delta t})
-
(\frac{\mathbf{p}_t-\mathbf{p}_{t-1}}{\Delta t})
\right\|_2,
\quad \Delta t = 0.1 \, s.
\end{aligned}
\end{equation}
}
Rather than acting as a generic smoothness prior, this term constrains the reconstructed dynamics to match the target dynamics, reducing frame-to-frame jitter that would otherwise be amplified downstream, since the \textit{forecaster} uses velocity and acceleration features derived from the \textit{projector} output.

The soft spatial-overlap term is defined as:

{
\begin{equation}
\begin{aligned}
\mathcal{L}_{\mathrm{overlap}}
=
1
-
\frac{1}{|\mathcal{S}|}
\sum_{\sigma \in \mathcal{S}}
\frac{1}{\sum_t m_t} \\
\sum_t
m_t 
\exp 
\left(
-
\frac{
\left\|
\hat{\mathbf{p}}_t-\mathbf{p}_t
\right\|_2^2
}{2\sigma^2}
\right),
\end{aligned}
\end{equation}
} 
where $\mathcal{S}$ is a set of Gaussian bandwidths. Unlike hard spatial constraints, Intersection over Union (IoU) style overlap losses, or map-based penalties, this term measures soft agreement directly in the BEV coordinate domain. Small bandwidths reward precise local alignment, while larger bandwidths preserve useful gradients for larger errors. 

Hence, the full \textit{projector} loss applied on normalized BEV image-plane values becomes: 

{
\begin{equation}
\mathcal{L}_{\mathrm{proj}}
=
\mathcal{L}_{\mathrm{traj}}
\left(
\hat{\mathbf{pos}}_{xy}^{\mathrm{BEV}},
\mathbf{pos}_{xy}^{\mathrm{BEV}}
\right)     
+
\mathcal{L}_{head_{size}}
+
\mathcal{L}_{head_{vel}},
\end{equation}
}

where $\hat{\mathbf{pos}}_{xy}$ represents the BEV position on the $x$-$y$ plane, and $\mathcal{L}_{\mathrm{head_{size}}}$ and $\mathcal{L}_{\mathrm{head_{vel}}}$ are $\mathcal{L}_{1}$ losses applied to size and velocity, respectively.

The \textit{forecaster} loss applied to metric BEV coordinates becomes:

\begin{equation}
\mathcal{L}_{\mathrm{forc}}
=
\mathcal{L}_{\mathrm{traj}}
\left(
\hat{\mathbf{P}}^{m}_{\mathrm{fut}},
\mathbf{P}^{\mathrm{m}}_{\mathrm{fut}}
\right).
\end{equation}

\section{Experimental Setup}
\label{sec:experimental_setup}

This section describes the adopted metrics, evaluation protocols, and implementation details for our two experimental settings. The first evaluates the full pipeline on VisDrone and R3ST++, while the second assesses performance on the external roadside benchmark TUMTraf-V2X.

\subsection{Evaluation Metrics}

Detection performance is reported in terms of Precision, Recall, mAP$_{50}$, and mAP$_{50:95}$. Cross-camera matching is evaluated at pairwise, frame-assignment, and tracklet-association levels using Accuracy, Precision, Recall, and F1 score when applicable. BEV projection and forecasting are evaluated with ADE and FDE in meters. Runtime is reported as per-stage latency, end-to-end latency, and FPS.

\subsection{Full Pipeline Evaluation on VisDrone and R3ST++}

\begin{table*}[!htbp]
\centering
\caption{Per-module training hyperparameters. The detector backbone is first trained on VisDrone with Stochastic Gradient Descent (SGD) and then fine-tuned (FT) on R3ST++, while all other modules use AdamW with gradient clipping at norm~1.0. Schedulers: W+L~=~linear warmup followed by linear decay. Plateau~=~ReduceLROnPlateau (factor~0.5, patience~5). W+C~=~linear warmup followed by cosine annealing. LR and ES denote learning rate and early stopping, respectively.}
\footnotesize
\setlength{\tabcolsep}{3pt}
\renewcommand{\arraystretch}{1.15}
\label{tab:hyperparams}
\begin{tabularx}{\textwidth}{c*{7}{>{\centering\arraybackslash}X}}
\toprule
\multirow{2}{*}{\textbf{Hyperparameter}} &
\multicolumn{2}{c}{\textbf{\textit{Detector}}} &
\multicolumn{2}{c}{\textbf{\textit{Detector} FT}} &
\multirow{2}{*}{\textbf{\textit{Matcher}}} &
\multirow{2}{*}{\textbf{\textit{Projector}}} &
\multirow{2}{*}{\textbf{\textit{Forecaster}}} \\
\cmidrule(lr){2-3}\cmidrule(lr){4-5}
&
\textbf{\textit{Backbone}} &
\textbf{\textit{Refiner}} &
\textbf{\textit{Backbone} FT} &
\textbf{\textit{Refiner} FT} &
&
&
\\
\midrule

Dataset
& VisDrone 
& VisDrone
& R3ST++ 
& R3ST++
& R3ST++
& R3ST++
& R3ST++ 
\\

Optimizer
& SGD 
& AdamW
& SGD 
& AdamW
& AdamW
& AdamW
& AdamW \\

Learning rate
& $10^{-2}$ 
& $5 \times 10^{-4}$
& $10^{-3}$ 
& $5 \times 10^{-4}$
& $10^{-3}$
& $10^{-3}$
& $5 \times 10^{-4}$ \\

Weight decay
& $5 \times 10^{-4}$ 
& $10^{-4}$
& $5 \times 10^{-4}$ 
& $10^{-4}$
& $10^{-4}$
& $5 \times 10^{-4}$
& $10^{-4}$ \\

LR scheduler
& W+L
& Plateau
& W+L
& Plateau
& Plateau
& W+C
& W+C \\

Final LR
& $10^{-4}$
& $10^{-6}$
& $10^{-4}$
& $10^{-6}$
& $10^{-6}$
& $10^{-7}$
& $10^{-5}$ \\

Max epochs
& 100
& 80
& 50
& 40
& 300
& 150
& 200 \\

ES patience
& 50
& 12
& 15
& 6
& 10
& 20
& 20 \\

Batch size
& 8
& 8
& 8
& 8
& 64
& 64
& 64 \\

\bottomrule
\end{tabularx}
\end{table*}

\subsubsection{Evaluation Protocols}
\label{ss:es_rep}

The \textit{detector} is first evaluated on the VisDrone-DET validation set after training on the corresponding training split. It is then fine-tuned on R3ST++ and evaluated as part of the full pipeline, together with the remaining stages, which are trained from scratch within each fold under an 8-fold cross-validation protocol over the 23 recording sessions of the Tianjin setting. Splits are generated at the recording-session level to avoid leakage between trajectory windows: seven folds follow a 15/5/3 train/validation/test partition, while the last fold follows a 16/5/2 partition, so that each session appears in the test set exactly once. The observed window and prediction horizon are fixed to $T_{\mathrm{obs}}=12$ and $T_{\mathrm{fut}}=12$ frames at 10~Hz, following the common setting used in literature for SIND-based evaluations. Valid-timestep masks are applied to both the observed and future sequences during training, so that losses and metrics are computed only over available timesteps, with indices ranging from 0 to 11. 

\subsubsection{Training and Implementation Details}
\label{sss_r3st_train_imp_det}

The training hyperparameters of our architecture are summarized in Table~\ref{tab:hyperparams}. All modules, with the exception of the \textit{backbone}, are optimized with AdamW, gradient clipping with norm 1.0, and early stopping on the validation loss.

The \textit{backbone} is a YOLO11l~\cite{khanam2024yolov11overviewkeyarchitectural} ($\approx$25.3M parameters, $\approx$25.5M the overall \textit{detector}) initialized from COCO-pretrained weights and trained on VisDrone at an input resolution of 640 pixels, with the default Ultralytics loss gains and augmentation pipeline (mosaic, RandAugment, HSV jittering, horizontal flipping, and random erasing). The \textit{refiner} is trained on detected crops with the \textit{backbone} frozen, using label smoothing of 0.05. For the \textit{tracker}, pairs below an IoU of $0.15$ are discarded, and unmatched tracks are retained for up to $10$ frames ($1\,\mathrm{s}$ at $10\,\mathrm{Hz}$).

Following the same procedure, the \textit{detector} is then fine-tuned on R3ST++ independently within each fold, using only the fold's training sessions, with unchanged optimizer, augmentation, and input resolution, but reduced learning rate and early-stopping patience. Frames are temporally split into training, validation, and test subsets using a 70/20/10 ratio, with a 300-frame gap between consecutive partitions to reduce leakage from temporally adjacent frames.

Regarding the \textit{matcher}, at inference time, the pairwise matching probabilities are first organized into a cost matrix $1-p_{ij,t}$ and resolved with the rectangular Hungarian algorithm \cite{ref12}. The resulting assignment is then made partial by discarding all candidate links whose probability is lower than $\eta=0.5$. Since the descriptor $r(\cdot)$ is built only from geometric cues and does not include the class label, the assignment remains purely geometric. Track-level correspondences are retained only when $C_{\min}=3$ and $\alpha_{\min}=0.30$ are satisfied.

\textit{Projector} and \textit{forecaster} loss weights are calibrated with the same protocol. Since AdamW makes the optimization approximately invariant to a global rescaling of the loss, only relative weights are searched. We fixed the ADE weight to 1.0 and sampled the remaining weights with a log-uniform Latin-Hypercube design in [0.2,5], centered on the uniform configuration. Each setting was evaluated on fold 0 using three different seeds, but with the same architecture, optimizer, scheduler, early-stopping protocol, and ranked with a validation objective combining normalized ADE, FDE, and velocity error. For both stages, the uniform configuration achieved the best score, and the best non-uniform setting was still 4.9~\% and 5.9~\% worse, respectively. For the soft-overlap term, the Gaussian bandwidth set $\mathcal{S}$ was fixed to (0.005, 0.02, 0.10) (normalized) for the \textit{projector} and to (0.05, 0.20, 1.0) m for the \textit{forecaster}. 

For the \textit{forecaster}, the egocentric-frame threshold was set to $\epsilon = 10^{-3}$ meters.

\subsection{External Roadside-Domain Evaluation on TUMTraf-V2X}

\subsubsection{Evaluation Protocols}
\label{sss:tumtraf_eval_protocol}

Based on the coverage of the sensing configuration described in Section~\ref{sec:dataset_tumtraf} and the occlusions caused by road-sign support infrastructure, we defined the BEV region of interest (ROI) as $x \in [0,70]$~meters and $y \in [-25,50]$~meters in the LiDAR coordinate space (see Fig. \ref{fig:roi_tumtraf}). We also normalized the trajectories in this region through a fixed global affine transform defined by this geometric ROI. LiDAR measurements are used as the temporal reference: camera observations are assigned to the nearest LiDAR timestamp, and trajectory windows are built on a $10$~Hz grid. The official train, validation, and test partitions are merged and re-segmented into $10$ temporally contiguous sequences, each lasting $10$~seconds. Since native 2D image annotations are not available, 2D reference boxes for the detection analysis are obtained by projecting each LiDAR cuboid into the corresponding camera and taking the axis-aligned box enclosing the projected cuboid vertices. These provide an approximate image-plane reference, which may be affected by cuboid-to-silhouette mismatches and residual camera-LiDAR misalignment. Cross-view ground-truth labels are obtained from persistent LiDAR track IDs projected into both cameras.  Missing per-frame states are filled through per-track interpolation within each sequence (linear for position and size, circular for orientation), filling only internal gaps of at most 8 frames so that reconstructed states never depend on frames from another split.

Given the limited amount of recorded data, TUMTraf-V2X is evaluated with a sequence-level 5-fold cross-validation protocol (5 training, 3 validation, and 2 test sequences per fold), with $T_{\mathrm{obs}}=T_{\mathrm{fut}}=12$ frames (1.2~s at 10~Hz) as on R3ST++.

\subsubsection{Training and Implementation Details}

The \textit{detector} uses the VisDrone-training checkpoint and is evaluated considering the shift-tolerant one-to-one association protocol described in Section~\ref{sss:tumtraf_eval_protocol}, using an IoU threshold of 0.3. The \textit{matcher} is trained from scratch within each fold. For the \textit{projector} and the \textit{forecaster}, we compare three training regimes under the same architecture: training from scratch, fine-tuning from the best R3ST++ checkpoint (Tianjin topology and the opposite-corner camera setup observing the intersection center), and fine-tuning from a lightly calibrated R3ST++ checkpoint, where the Tianjin topology and trajectories are kept but the cameras are configured according to the TUMTraf-V2X sensing geometry. Learning rates are set to $5 \times 10^{-4}$ for from-scratch training and $1 \times 10^{-4}$ for fine-tuning.

\begin{table}[t]
\centering
\newcolumntype{C}[1]{>{\centering\arraybackslash}p{#1}}
\renewcommand{\arraystretch}{1.05}
\begin{minipage}[c]{0.48\textwidth}
\centering
\caption{Detection performance against state-of-the-art one-stage detectors on the VisDrone-DET validation set at 640~px input.}
\label{tab:visdrone_sota}
\small
\setlength{\tabcolsep}{2pt}
\begin{tabular}{@{}
p{0.40\linewidth}
C{0.28\linewidth}
C{0.28\linewidth}
@{}}
\toprule
\textbf{Method} & \textbf{mAP$_{50}$} & \textbf{mAP$_{50:95}$} \\
\midrule
MASF-YOLO-S \cite{masf2025}                & \textbf{0.492} & \textbf{0.329} \\
\textbf{Our \textit{backbone}}             & 0.454 & 0.273 \\
\textbf{\textit{Our backbone} + \textit{refiner}}  & 0.487 & 0.312 \\
MSFE-YOLO-L \cite{msfe2024}                & 0.468 & 0.290 \\
LRDS-YOLO \cite{lrds2025}                  & 0.436 & 0.266 \\
YOLOv9-M \cite{wang2024yolov9}             & 0.420 & 0.252 \\
YOLO-MARS \cite{mars2025}                  & 0.409 & 0.234 \\
YOLOv8-M \cite{jocher2023yolov8}           & 0.407 & 0.246 \\
YOLOv10-M \cite{wang2024yolov10}           & 0.405 & 0.245 \\
DASSF \cite{dassf2024}                     & 0.396 & 0.235 \\
Gold-YOLO-S \cite{wang2023goldyolo}        & 0.386 & 0.233 \\
\bottomrule
\end{tabular}
\end{minipage}\hfill
\begin{minipage}[c]{0.48\textwidth}
\centering
\caption{Fine-tuned \textit{detector} performance on the R3ST++ fold 0.}
\label{tab:r3st_after_ft_per_class}
\small
\setlength{\tabcolsep}{3pt}
\begin{tabular}{@{}lcccc@{}}
\toprule
\textbf{Class} & \textbf{Precision} & \textbf{Recall} & \textbf{mAP$_{50}$} & \textbf{mAP$_{50:95}$} \\
\midrule
Car                & 0.990 & 0.974 & 0.994 & 0.976 \\
Bus                & 0.985 & 0.971 & 0.984 & 0.972 \\
Truck              & 0.987 & 0.973 & 0.986 & 0.974 \\
Bicycle            & 0.990 & 0.939 & 0.993 & 0.917 \\
Tricycle           & 0.987 & 0.938 & 0.990 & 0.913 \\
Motorcycle         & 0.989 & 0.933 & 0.986 & 0.908 \\
Pedestrian         & 0.962 & 0.913 & 0.987 & 0.918 \\
\midrule
\textbf{All}       & \textbf{0.984} & \textbf{0.949} & \textbf{0.989} & \textbf{0.940} \\
\bottomrule
\end{tabular}
\end{minipage}
\end{table}

\section{Experimental Results}
\label{sec:experimental_results}

\subsection{Full Pipeline Evaluation on VisDrone and R3ST++}
\label{subsec:exp_r3st}

\begin{figure*}[t]
\centering
\includegraphics[width=0.32\textwidth]{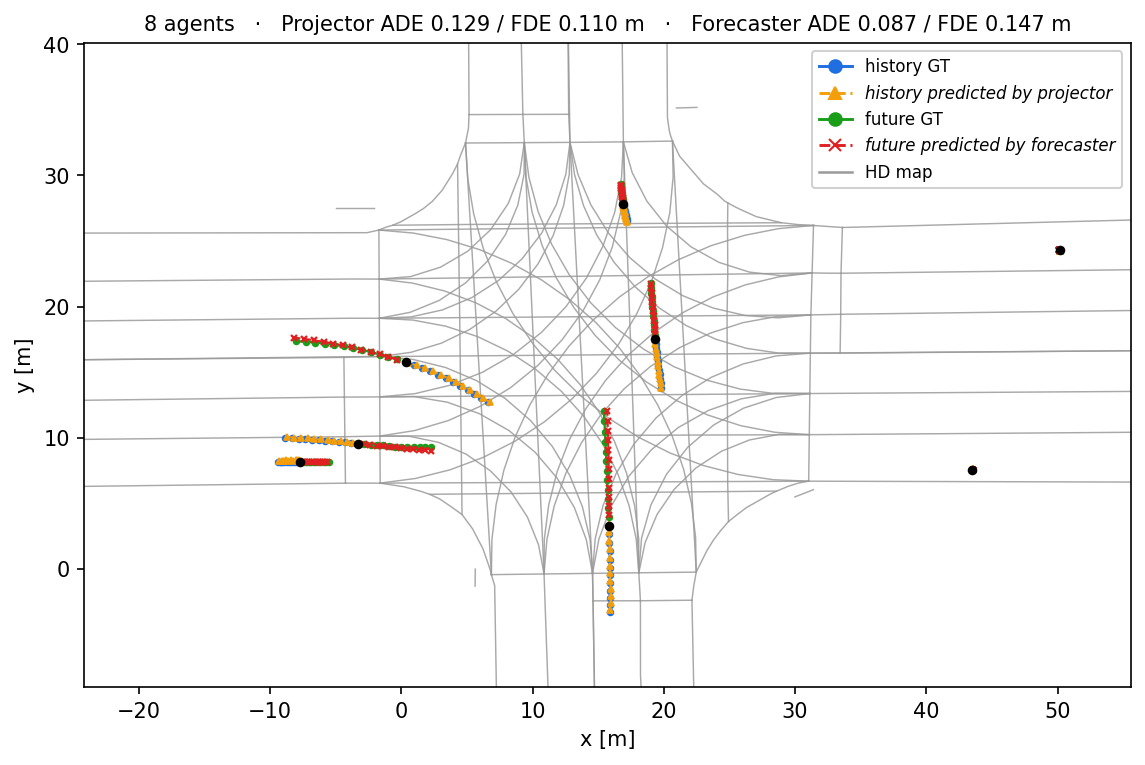}
\includegraphics[width=0.32\textwidth]{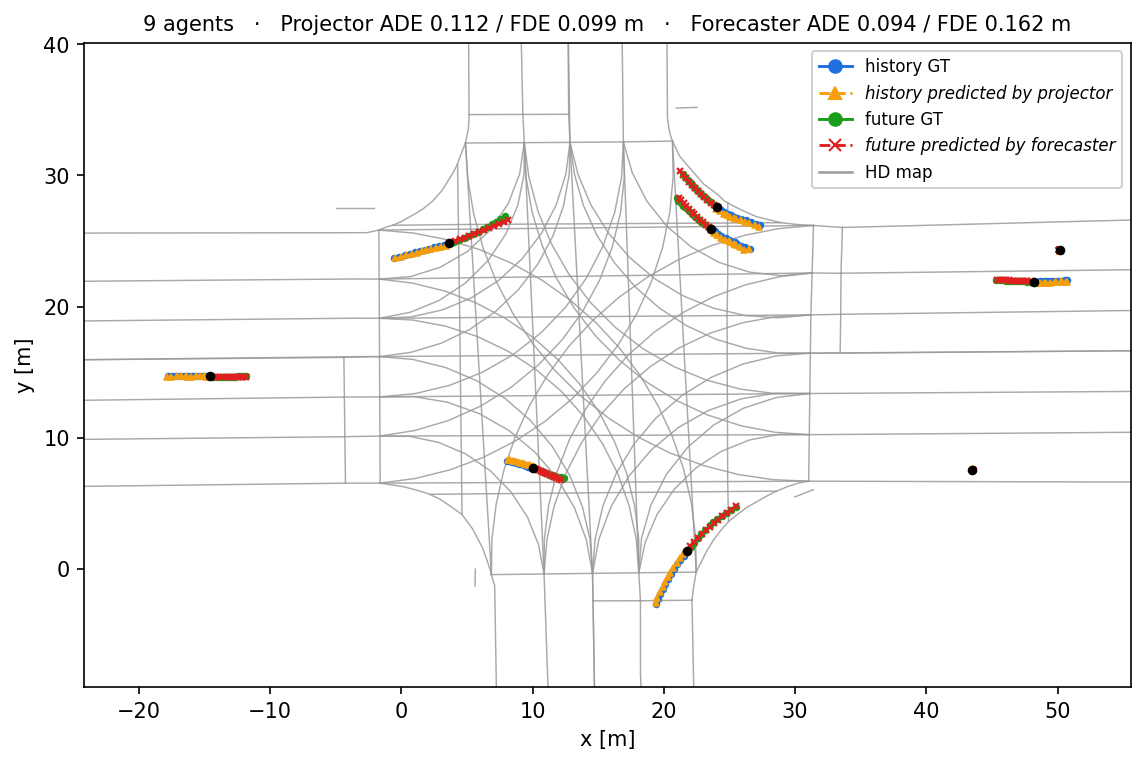}
\includegraphics[width=0.32\textwidth]{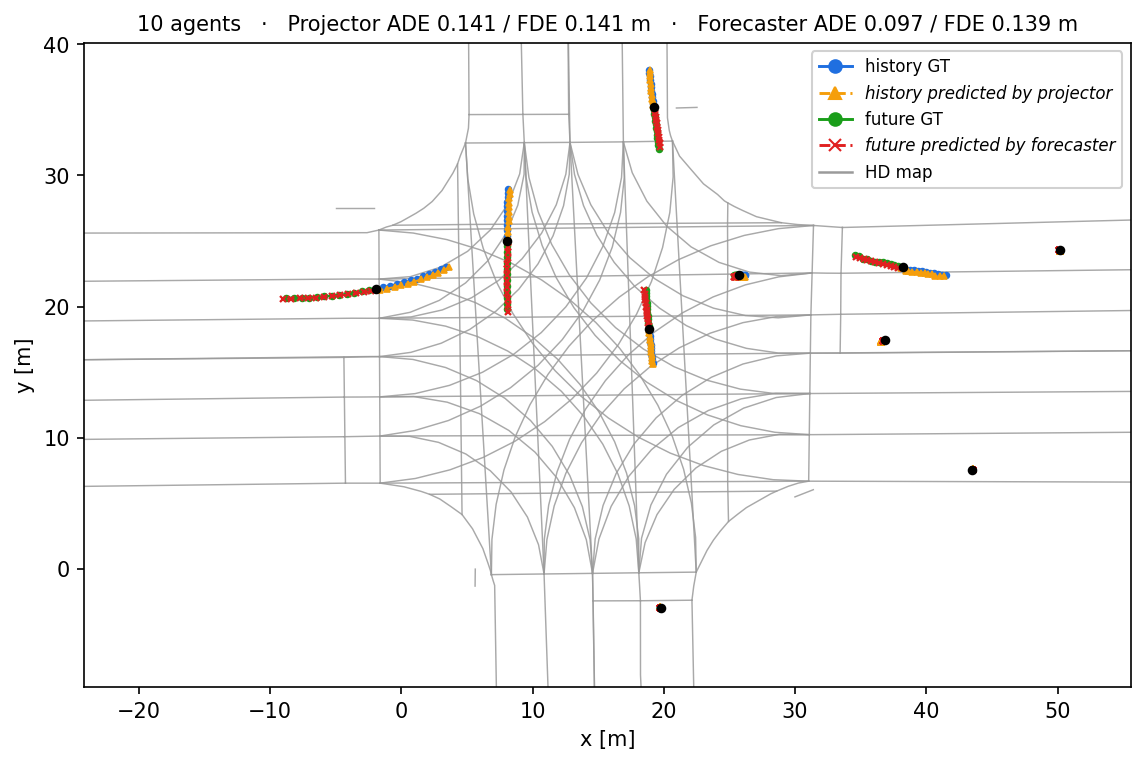}
\caption{Qualitative OCTR2A results with average errors per scene on R3ST++ with 12 observed and 12 forecasted frames compared with BEV ground truth from SIND.}
\label{fig:R3ST_qualitative_results}
\end{figure*}

\begin{table}[t]
\centering
\caption{Cross-camera matching performance on R3ST++ at pairwise, one-to-one frame-level assignment, and temporal tracklet levels (mean~$\pm$~std over the 8 folds).}
\label{tab:stage2_results}
{
\setlength{\columnwidth}{0.85\textwidth}
\small
\setlength{\tabcolsep}{1.5pt}
\renewcommand{\arraystretch}{1.05}
\newcolumntype{C}[1]{>{\centering\arraybackslash}p{#1}}
\begin{tabular}{@{}
p{0.15\columnwidth}
C{0.19\columnwidth}
C{0.19\columnwidth}
C{0.19\columnwidth}
C{0.19\columnwidth}
@{}}
\toprule
\textbf{Level} & \textbf{Accuracy} & \textbf{Precision} & \textbf{Recall} & \textbf{F1} \\
\midrule
Pairwise cls.
& \makecell{$0.993$ $\pm 0.004$}
& \makecell{$0.901$ $\pm 0.035$}
& \makecell{$0.999$ $\pm 0.000$}
& \makecell{$0.948$ $\pm 0.019$} \\

Frame-level
& \makecell{$0.999$ $\pm 0.001$}
& \makecell{$0.997$ $\pm 0.003$}
& \makecell{$0.998$ $\pm 0.002$}
& \makecell{$0.999$ $\pm 0.001$} \\

Tracklet assoc.
& -
& \makecell{$1.000$ $\pm 0.000$}
& \makecell{$1.000$ $\pm 0.000$}
& \makecell{$1.000$ $\pm 0.000$} \\
\bottomrule
\end{tabular}\par
}

\vspace{1.5em}

\caption{BEV reconstruction and forecasting errors of OCTR2A on R3ST++, in meters (mean~$\pm$~std over the 8 folds).}
\label{tab:stage34_results}
\begin{tabular}[\columnwidth]{ccc}
\toprule
\textbf{Stage} & \textbf{ADE (m)} & \textbf{FDE (m)} \\
\midrule
Stage 3: \textit{projector}
& $0.1440 \pm 0.0096$
& $0.1401 \pm 0.0111$ \\

Stage 4: \textit{forecaster}
& $0.1537 \pm 0.0106$
& $0.2408 \pm 0.0154$ \\
\bottomrule
\end{tabular}

\vspace{1.5em}

\caption{Runtime (in ms) and FPS performance of OCTR2A on GPU and CPU, for an increasing number of agents per frame. S.1: \textit{detector} over two cameras. S.2: \textit{matcher}, S.3: \textit{projector}, and S.4: \textit{forecaster}.}
\label{tab:runtime_results_n_agents}
\setlength{\columnwidth}{1.0\textwidth}
\small
\setlength{\tabcolsep}{2pt}
\renewcommand{\arraystretch}{1.05}
\newcolumntype{C}[1]{>{\centering\arraybackslash}p{#1}}
\begin{tabular}{@{}
C{0.10\columnwidth}
C{0.33\columnwidth}
C{0.08\columnwidth}
C{0.33\columnwidth}
C{0.08\columnwidth}
@{}}
\toprule
\multirow{3}{*}{\textbf{Agents}} &
\multicolumn{2}{c}{\textbf{GPU (NVIDIA GeForce GTX 1080 Ti)}} &
\multicolumn{2}{c}{\textbf{CPU (Intel(R) Core(TM) i9-7920X)}} \\
\cmidrule(lr){2-3}
\cmidrule(lr){4-5}
&
\textbf{S.1 / S.2 / S.3 / S.4 / Full (ms)} &
\textbf{FPS} &
\textbf{S.1 / S.2 / S.3 / S.4 / Full (ms)} &
\textbf{FPS} \\
\midrule
8 &
31.5 / 0.7 / 1.5 / 1.1 / 35.2 &
28 &
368.3 / 0.7 / 2.4 / 2.1 / 373.7 &
3 \\
16 &
31.5 / 1.2 / 1.7 / 1.2 / 35.5 &
28 &
368.3 / 1.5 / 2.8 / 2.3 / 375.2 &
3 \\
32 &
31.5 / 3.7 / 1.7 / 1.2 / 37.8 &
26 &
368.3 / 4.2 / 3.5 / 2.8 / 379.2 &
3 \\
64 &
31.5 / 12.7 / 1.7 / 1.3 / 47.3 &
21 &
368.3 / 14.8 / 4.5 / 3.6 / 393.1 &
3 \\
\bottomrule
\end{tabular}

\end{table}

On the VisDrone-DET validation set (Table~\ref{tab:visdrone_sota}), the \textit{refiner} improves the \textit{backbone} performance from 0.454~/~0.273 to 0.487~/~0.312 mAP$_{50}$~/~mAP$_{50:95}$ without altering the predicted box geometry. Overall, the \textit{detector} ranks second among the considered one-stage detectors, indicating strong generalization capability and supporting the contribution of semantic consistency to detection performance. The class-wise results in Table~\ref{tab:r3st_after_ft_per_class} illustrate the \textit{detector}'s performance after fine-tuning on fold~0 of R3ST++. The \textit{detector} reaches 0.989 mAP$_{50}$ overall, with the highest localization accuracy on large motorized classes and lower recall on vulnerable road users, consistent with their stronger scale variation and partial visibility in oblique roadside views.

As shown in Table~\ref{tab:stage2_results}, the \textit{matcher} reaches 0.948 F1 at the pairwise level with near-saturated recall, denoting a classifier that avoids missed correspondences. The one-to-one frame-level assignment removes most inconsistent links (0.999 F1), and tracklet-level association attains perfect precision and recall across folds. Geometric pair scoring, bipartite assignment, and temporal consistency are therefore sufficient to recover stable cross-view identities.

The \textit{projector} reconstructs observed BEV histories with 0.1440~m ADE and 0.1401~m FDE (see Table~\ref{tab:stage34_results}). The closeness of these two values indicates a stable error that does not accumulate toward the last observed frame, which is essential considering that the \textit{forecaster} consumes dynamics recomputed from this history. The latter predicts the 12-frame horizon with an ADE of 0.1537~m and an FDE of 0.2408~m (see Table~\ref{tab:stage34_results}). Although FDE is higher than ADE due to increasing uncertainty along the horizon, these values confirm accurate forecasting from reconstructed projected trajectories and show that OCTR2A can map synchronized RGB inputs to metric BEV predictions with low error, as also qualitatively illustrated in Fig.~\ref{fig:R3ST_qualitative_results}.

\subsubsection{Inference Latency and Throughput}

On an NVIDIA GTX 1080~Ti, the full pipeline runs at 35.2--47.3~ms per frame for 8--64 agents (28$\rightarrow$21 FPS), well within the 100~ms budget imposed by the 10~Hz data (see Table~\ref{tab:runtime_results_n_agents}). Detection is the dominant latency contributor (31.5~ms for both views), while the remaining stages add only 3.7--15.8~ms. On Intel Core i9-7920X CPU, the same configuration runs at approximately 3~FPS, confirming that GPU execution is required for real-time RGB-to-BEV processing.

\begin{table}[t]
\centering
\caption{\textit{Projector} vs. calibrated geometric baselines errors in meters on R3ST++ (mean~$\pm$~std over the 8 folds). Coverage is the fraction of trajectory windows where a method is applicable.}
\label{tab:geomBaselinesAligned}
\setlength{\columnwidth}{0.92\textwidth}
\small
\setlength{\tabcolsep}{2pt}
\renewcommand{\arraystretch}{1.05}
\newcolumntype{C}[1]{>{\centering\arraybackslash}p{#1}}
\begin{tabular}{@{}
p{0.44\columnwidth}
C{0.17\columnwidth}
C{0.17\columnwidth}
C{0.12\columnwidth}
@{}}
\toprule
\textbf{Method} & \textbf{ADE (m)} & \textbf{Median ADE} & \textbf{Coverage} \\
\midrule
\textbf{\textit{projector}} & \textbf{0.144 $\pm$ 0.010} & \textbf{0.125 $\pm$ 0.006} & 100~\% \\
Two-view DLT + homography fallback & 1.129 $\pm$ 0.063 & 0.739 $\pm$ 0.093 & 100~\% \\
Ray–plane homography & 1.284 $\pm$ 0.065 & 0.899 $\pm$ 0.060 & 100~\% \\
\midrule
\textbf{\textit{projector} on single-view windows} & \textbf{0.157 $\pm$ 0.024} & \textbf{0.139 $\pm$ 0.024} & 59.1~\% \\
Ray–plane homography & 1.577 $\pm$ 0.052 & 1.368 $\pm$ 0.123 & 59.1~\% \\
\midrule
\textbf{\textit{projector} on two-view windows} & \textbf{0.126 $\pm$ 0.025} & 0.115 $\pm$ 0.018 & 40.9~\% \\
Two-view DLT triangulation & 0.388 $\pm$ 0.186 & \textbf{0.080 $\pm$ 0.008} & 40.9~\% \\
Ray–plane homography & 0.808 $\pm$ 0.187 & 0.547 $\pm$ 0.021 & 40.9~\% \\
\bottomrule
\end{tabular}
\end{table}

\subsection{Projection and Forecasting Module Evaluation}

\subsubsection{BEV \textit{Projector} vs. Geometric Baselines on R3ST++}

We evaluate the BEV \textit{projector} against geometric baselines using exact simulator calibration, identical bounding-box inputs, and the same 8-fold protocol (Table~\ref{tab:geomBaselinesAligned}). We use a ray-plane estimator, equivalent to applying a ground-plane homography, that maps each bounding-box bottom center to its BEV position. When both cameras observe an agent, we average the two estimates, while under single-view visibility, we use the available estimate. As a distinct estimator, two-view Direct Linear Transform (DLT) triangulates the bounding-box centers and retains the horizontal coordinates of the reconstructed 3D point. The full-coverage hybrid baseline uses DLT under dual visibility and falls back to the ray-plane estimator otherwise.

Under full coverage, the \textit{projector} obtains 0.144~m ADE against 1.129~m for the best geometric baseline (7.8$\times$ lower), with the gain also evident in the median (0.125~m vs. 0.739~m), indicating a consistently more accurate typical reconstruction rather than only fewer large failures.

The advantage persists across visibility regimes: on single-view windows the \textit{projector} reaches 0.157~m ADE against 1.577~m for single-view homography, while on two-view windows it reduces the error to 0.126~m against 0.388~m for triangulation (3.1$\times$ improvement). Although triangulation attains a lower median in favorable configurations, its larger mean and variance reveal unstable behavior for large agents and geometrically sensitive viewpoints. The learned \textit{projector} therefore provides more robust BEV reconstruction than explicit calibrated mappings in both regimes.

\begin{table}[t]
\centering
\caption{Forecasting comparison on SIND at 10~Hz, for the 12-12 and 12-18 history/future settings (meters, mean~$\pm$~std over the 8 folds). Subscript~$k$ is the number of predicted trajectories. S and M mark the use of traffic signal and HD-map information.}
\label{tab:SIND_i2xtraj_10hz}
\setlength{\columnwidth}{0.9\textwidth}
\small
\setlength{\tabcolsep}{1.2pt}
\renewcommand{\arraystretch}{1.05}
\newcolumntype{C}[1]{>{\centering\arraybackslash}p{#1}}
\begin{tabular}{@{}
p{0.22\columnwidth}
C{0.15\columnwidth}
C{0.15\columnwidth}
C{0.15\columnwidth}
C{0.15\columnwidth}
C{0.05\columnwidth}
C{0.05\columnwidth}
@{}}
\toprule
\multirow{3}{*}{\textbf{Method}} &
\multicolumn{2}{c}{\textbf{12-12}} &
\multicolumn{2}{c}{\textbf{12-18}} &
\multirow{3}{*}{\textbf{S}} &
\multirow{3}{*}{\textbf{M}} \\
\cmidrule(lr){2-3}
\cmidrule(lr){4-5}
& \textbf{ADE} & \textbf{FDE} & \textbf{ADE} & \textbf{FDE} & & \\
\midrule
S-GAN$_{6}$~\cite{ref27} & 1.32 & 2.46 & 1.53 & 2.95 & $\times$ & $\times$ \\
S-LSTM$_{1}$~\cite{ref26} & 0.87 & 1.60 & 0.96 & 1.78 & $\times$ & $\times$ \\
Trajectron++$_{\mathrm{6}}$~\cite{ref30} & 0.37 & 0.93 & 0.70 & 1.91 & $\times$ & $\times$ \\
FJMP$_{6}$~\cite{ref34} & 0.27 & 0.69 & 0.41 & 1.13 & $\times$ & $\times$ \\
HPNet$_{6}$~\cite{ref38} & 0.09 & 0.21 & 0.11 & 0.25 & $\times$ & $\checkmark$ \\
I2XTraj$_{6}$~\cite{ref6} & 0.09 & 0.12 & 0.10 & 0.21 & $\checkmark$ & $\checkmark$ \\
\textbf{Our \textit{forecaster}$_{1}$} & \textbf{0.03 $\pm$ 0.00} & \textbf{0.07 $\pm$ 0.01} & \textbf{0.06 $\pm$ 0.00} & \textbf{0.16 $\pm$ 0.01} & $\times$ & $\times$ \\
\bottomrule
\end{tabular}
\end{table}

\begin{table}[t]
\centering
\caption{Loss-term ablation for the BEV \textit{projector} and \textit{forecaster} in meters (mean~$\pm$~std over the 8 folds).}
\label{tab:loss_ablation}
\small
\setlength{\tabcolsep}{2.5pt}
\renewcommand{\arraystretch}{1.08}
{%
\begin{tabular}{@{}lcccc@{}}
\toprule
\multicolumn{5}{c}{\textbf{\textit{Projector}}} \\
\midrule
\textbf{Configuration} & \textbf{ADE} & \textbf{$\Delta$ADE} & \textbf{FDE} & \textbf{$\Delta$FDE} \\
\midrule
Uniform weights & $\mathbf{0.145{\pm}0.014}$ & - & $\mathbf{0.149{\pm}0.013}$ & - \\
w/o $\mathcal{L}_{\mathrm{ADE}}$ & $0.165{\pm}0.016$ & $+13.8~\%$ & $0.152{\pm}0.017$ & $+2.0~\%$ \\
w/o $\mathcal{L}_{\mathrm{FDE}}$ & $0.164{\pm}0.010$ & $+13.5~\%$ & $0.174{\pm}0.008$ & $+17.0~\%$ \\
w/o $\mathcal{L}_{\mathrm{VEL}}$ & $0.147{\pm}0.018$ & $+1.5~\%$ & $0.155{\pm}0.010$ & $+3.9~\%$ \\
w/o $\mathcal{L}_{\mathrm{overlap}}$ & $0.270{\pm}0.018$ & $+86.9~\%$ & $0.198{\pm}0.025$ & $+33.2~\%$ \\
w/o $\mathcal{L}_{\mathrm{head_{size}}}$ & $0.150{\pm}0.014$ & $+3.6~\%$ & $0.155{\pm}0.015$ & $+4.1~\%$ \\
w/o $\mathcal{L}_{\mathrm{head_{vel}}}$ & $0.161{\pm}0.021$ & $+11.5~\%$ & $0.161{\pm}0.016$ & $+7.9~\%$ \\
w/o $\mathcal{L}_{\mathrm{VEL}}$ and $\mathcal{L}_{\mathrm{head_{vel}}}$ & $0.152{\pm}0.023$ & $+5.1~\%$ & $0.160{\pm}0.022$ & $+7.6~\%$ \\
\midrule
\multicolumn{5}{c}{\textbf{\textit{Forecaster}}} \\
\midrule
\textbf{Configuration} & \textbf{ADE} & \textbf{$\Delta$ADE} & \textbf{FDE} & \textbf{$\Delta$FDE} \\
\midrule
Uniform weights & $\mathbf{0.147{\pm}0.012}$ & - & $\mathbf{0.230{\pm}0.016}$ & - \\
w/o $\mathcal{L}_{\mathrm{ADE}}$ & $0.151{\pm}0.012$ & $+2.7~\%$ & $0.232{\pm}0.011$ & $+0.9~\%$ \\
w/o $\mathcal{L}_{\mathrm{FDE}}$ & $0.150{\pm}0.013$ & $+1.7~\%$ & $0.236{\pm}0.017$ & $+2.6~\%$ \\
w/o $\mathcal{L}_{\mathrm{VEL}}$ & $0.148{\pm}0.012$ & $+0.4~\%$ & $0.231{\pm}0.016$ & $+0.3~\%$ \\
w/o $\mathcal{L}_{\mathrm{overlap}}$ & $0.159{\pm}0.012$ & $+8.2~\%$ & $0.242{\pm}0.016$ & $+5.2~\%$ \\
\bottomrule
\end{tabular}%
}
\end{table}

\subsubsection{BEV \textit{Forecaster} vs. State-of-the-Art Methods on SIND}

Table~\ref{tab:SIND_i2xtraj_10hz} shows the performance comparison between the \textit{forecaster} and recent SOTA methods on SIND at 10~Hz, using 12 observed frames, and both 12-frame and 18-frame prediction horizons. In the 12-12 setting, the \textit{forecaster} achieves 0.03~m ADE and 0.07~m FDE, improving over the strongest competitor by 66.7\% and 41.7\%, respectively. In the 12-18 setting, it obtains 0.06~m ADE and 0.16~m FDE, with relative gains of 40.0\% and 23.8\% over the best alternative. These results are obtained with a single predicted trajectory and without using traffic-signal states, HD map, or other scene-context inputs, whereas several competing methods rely on multiple hypotheses or additional infrastructure information. The comparison therefore suggests that the egocentric GRU formulation captures the short-term dynamics of intersection agents effectively from BEV histories alone, while preserving strong accuracy also over the longer forecasting horizon.

\begin{table}[t]
\centering
\caption{System-level ablation of the \textit{forecaster} training strategy under cascaded inference (C.I.). standalone training of the \textit{projector} and \textit{forecaster} (S.T.), \textit{forecaster} fine-tuning on projected histories (F.T.), and cascade training (C.T.). Results in meters (mean~$\pm$~std over the 8 folds).}
\label{tab:system_level_ablation_8fold}
{
\setlength{\columnwidth}{0.85\textwidth}
\small
\setlength{\tabcolsep}{1.6pt}
\renewcommand{\arraystretch}{1.05}
\newcolumntype{C}[1]{>{\centering\arraybackslash}p{#1}}
\begin{tabular}{@{}
p{0.26\columnwidth}
C{0.22\columnwidth}
C{0.22\columnwidth}
C{0.22\columnwidth}
@{}}
\toprule
\textbf{Metric} & \textbf{S.T. + C.I.} & \textbf{S.T. + F.T. + C.I.} & \textbf{C.T. + C.I.} \\
\midrule
\textit{forecaster} ADE (m)
& $0.241 \pm 0.029$
& $0.207 \pm 0.024$
& $\mathbf{0.154 \pm 0.011}$ \\

\textit{forecaster} FDE (m)
& $0.399 \pm 0.051$
& $0.377 \pm 0.044$
& $\mathbf{0.240 \pm 0.015}$ \\
\bottomrule
\end{tabular}\par
}

\vspace{1.5em}

\caption{Fold-0 ablation of the \textit{forecaster} input: BEV history source (ground truth, GT, vs. \textit{projector}) and velocity input (finite-difference, FD, vs. predicted). ADE/FDE in meters.}
\label{tab:controlled_ablation_fold0}
{
\setlength{\columnwidth}{0.85\textwidth}
\small
\setlength{\tabcolsep}{1.4pt}
\renewcommand{\arraystretch}{1.08}
\newcolumntype{C}[1]{>{\centering\arraybackslash}p{#1}}
\begin{tabular}{@{}
p{0.23\columnwidth}
p{0.22\columnwidth}
p{0.22\columnwidth}
C{0.12\columnwidth}
C{0.12\columnwidth}
@{}}
\toprule
\textbf{Training variant} & \textbf{BEV history} & \textbf{Velocity input} & \textbf{ADE} & \textbf{FDE} \\
\midrule
Standalone
& GT history
& FD velocity
& $0.168$
& $0.295$ \\

Cascaded
& \textit{projector} output
& FD velocity
& $0.112$ 
& $0.203$  \\

Cascaded + vel. head
& \textit{projector} output
& Pred. velocity
& $\mathbf{0.105}$
& $\mathbf{0.194}$ \\
\bottomrule
\end{tabular}\par
}
\end{table}

\subsubsection{Loss-Term and Input-Source Ablation Study}

Table~\ref{tab:loss_ablation} reports single-term removals (one weight zeroed, the rest set to one) over all folds and three seeds, plus a leave-two-out on the \textit{projector}'s velocity terms. The results show that the soft-overlap term is by far the most influential for the \textit{projector} (+86.9\% ADE, +33.2\% FDE when removed) and remains the dominant term for the \textit{forecaster} (+8.2\%/+5.2\%), while the ADE and FDE terms mainly affect the metric they directly supervise. The two velocity-related \textit{projector} terms are partially redundant (removing both degrades ADE by only 5.1\%), and the \textit{forecaster} is largely insensitive to the velocity-consistency term, likely because its velocity-increment decoder already encourages temporally coherent predictions. The size head in the \textit{projector} also contributes to the final reconstruction quality, although its impact is smaller. Table~\ref{tab:system_level_ablation_8fold} evaluates how the \textit{forecaster} output changes as its training input is progressively aligned with the projected histories used during cascaded inference. Standalone training on ground-truth histories yields 0.241/0.399~m ADE/FDE, then fine-tuning on projected histories improves this to 0.207/0.377~m, whereas full cascade training achieves 0.154/0.240~m, resulting in a 36.1\%/39.8\% gain over standalone training. Table~\ref{tab:controlled_ablation_fold0} further shows on fold~0 that most of this gain comes from matching the training and inference input distributions (0.168/0.295~m with ground-truth histories vs. 0.112/0.203~m with projected ones), with the \textit{projector} velocity head providing an additional improvement to 0.105/0.194~m over finite-difference velocities.

\begin{figure}[t]
\centering
\includegraphics[width=0.6\textwidth]{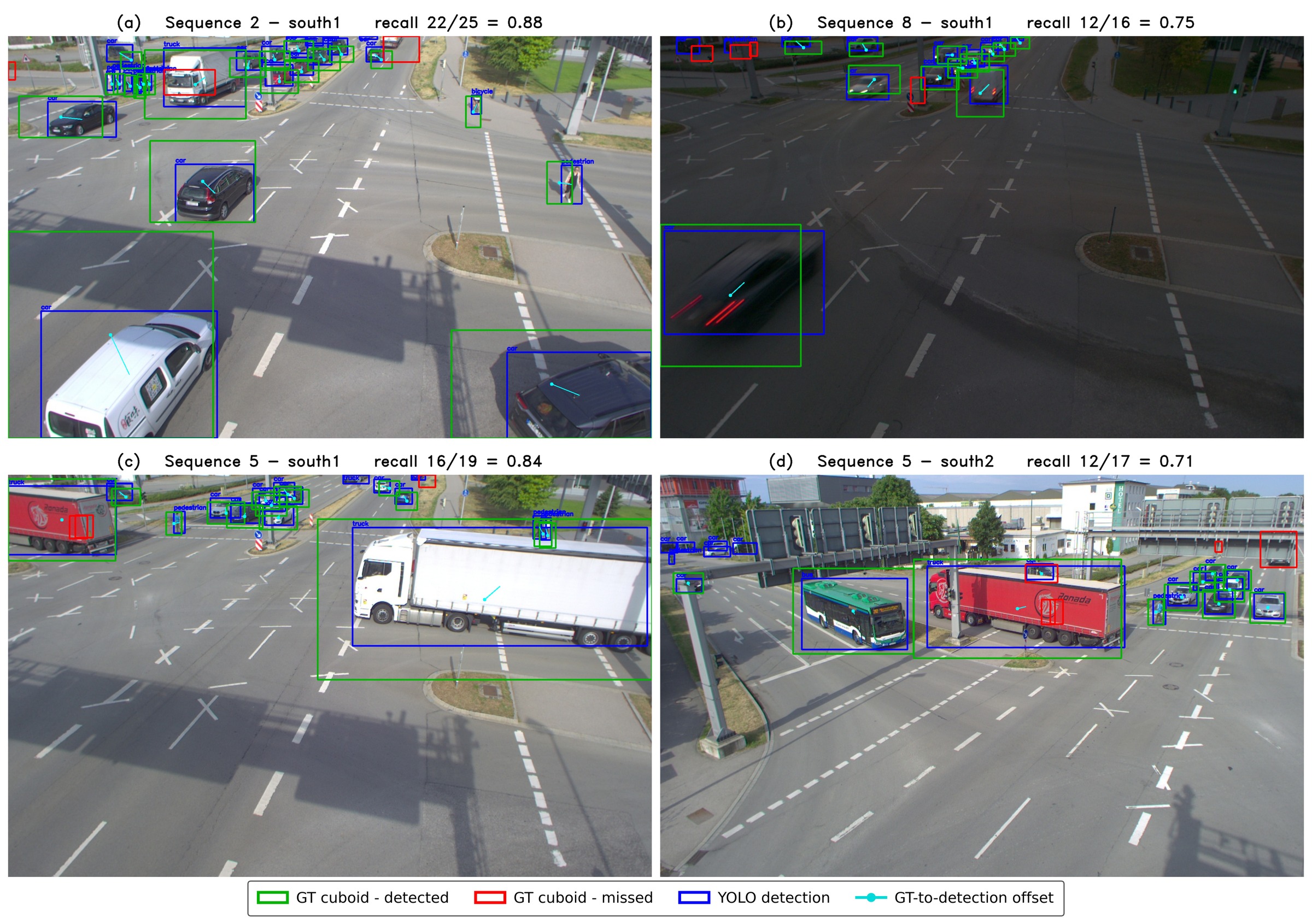}
\caption{Qualitative detection examples on TUMTraf-V2X. Green boxes denote matched projected GT-LiDAR cuboids, red boxes missed projected GT-LiDAR cuboids, blue boxes \textit{detector} detections, and cyan segments the GT-to-detection offset, highlighting visibility and cuboid-to-silhouette mismatches.}
\label{fig:stage1_gt_comparison_tumtraf}
\end{figure}

\begin{table}[t]
\centering
\caption{\textit{Detector} results on TUMTraf-V2X with shift-tolerant one-to-one association considering projected 2D LiDAR reference boxes.}
\label{tab:tumtraf_detection}
\setlength{\columnwidth}{0.7\textwidth}
\small
\setlength{\tabcolsep}{2.4pt}
\renewcommand{\arraystretch}{1.12}
\newcolumntype{C}[1]{>{\centering\arraybackslash}p{#1}}
\begin{tabularx}{\columnwidth}{@{}
>{\raggedright\arraybackslash}X
C{0.18\columnwidth}
C{0.15\columnwidth}
C{0.18\columnwidth}
C{0.12\columnwidth}
@{}}
\toprule
\textbf{Class} & \textbf{Support} & \textbf{Recall} & \textbf{Precision} & \textbf{F1} \\
\midrule

Car
& 13,507
& 0.802
& 0.646
& 0.716 \\

Truck
& 1,638
& 0.673
& 0.634
& 0.653 \\

Motorcycle
& 425
& 0.586
& 0.331
& 0.423 \\

Bus
& 532
& 0.962
& 0.347
& 0.510 \\

Pedestrian
& 6,989
& 0.694
& 0.577
& 0.630 \\

Bicycle
& 328
& 0.704
& 0.583
& 0.638 \\

\midrule

\textbf{All}
& \textbf{23,419}
& \textbf{0.759}
& \textbf{0.602}
& \textbf{0.671} \\

\bottomrule
\end{tabularx}
\end{table}

\begin{table}[!htbp]
\centering
\caption{Performance on TUMTraf-V2X by stage, input source, and training mode, reported as mean $\pm$ standard deviation over the 5 folds. FT denotes models fine-tuned from R3ST++ checkpoints, whereas from-scratch models are trained exclusively on TUMTraf-V2X.}
\label{tab:tumtraf_stage_results}
\setlength{\columnwidth}{0.85\textwidth}
\small
\setlength{\tabcolsep}{1.5pt}
\renewcommand{\arraystretch}{1.05}
\newcolumntype{L}[1]{>{\raggedright\arraybackslash}p{#1}}
\newcolumntype{C}[1]{>{\centering\arraybackslash}p{#1}}
\newcolumntype{Y}{>{\raggedright\arraybackslash}X}
\begin{tabularx}{\columnwidth}{@{}
L{0.145\columnwidth}
Y
C{0.2\columnwidth}
C{0.15\columnwidth}
C{0.22\columnwidth}
@{}}
\toprule
\textbf{Stage} & \textbf{Input} & \textbf{Training Mode} & \textbf{Metric} & \textbf{Result} \\
\midrule

\textit{Matcher}
& GT cuboids
& --
& F1
& $\mathbf{0.991 \pm 0.011}$ \\

\cmidrule(lr){2-5}

\textit{Matcher}
& \textit{Detector} outputs
& --
& F1
& $0.936 \pm 0.058$ \\

\midrule

\multirow[c]{6}{*}{\textit{Projector}}
& \multirow[c]{6}{=}{GT association}
& \multirow[c]{2}{=}{From-scratch}
& ADE (m)
& $2.063 \pm 0.516$ \\
&
&
& FDE (m)
& $2.218 \pm 0.532$ \\

\cmidrule(lr){3-5}

&
& \multirow[c]{2}{=}{FT - Tianjin sensing setup}
& ADE (m)
& $1.672 \pm 0.264$ \\
&
&
& FDE (m)
& $1.710 \pm 0.254$ \\

\cmidrule(lr){3-5}

&
& \multirow[c]{2}{=}{FT - TUMTraf sensing setup}
& ADE (m)
& $\mathbf{0.978 \pm 0.223}$ \\
&
&
& FDE (m)
& $\mathbf{1.038 \pm 0.220}$ \\

\cmidrule(lr){2-5}

\multirow[c]{6}{*}{\textit{Projector}}
& \multirow[c]{6}{=}{\textit{Matcher} output}
& \multirow[c]{2}{=}{From-scratch}
& ADE (m)
& $2.079 \pm 0.505$ \\
&
&
& FDE (m)
& $2.243 \pm 0.515$ \\

\cmidrule(lr){3-5}

&
& \multirow[c]{2}{=}{FT - Tianjin sensing setup}
& ADE (m)
& $1.659 \pm 0.276$ \\
&
&
& FDE (m)
& $1.699 \pm 0.265$ \\

\cmidrule(lr){3-5}

&
& \multirow[c]{2}{=}{FT - TUMTraf sensing setup}
& ADE (m)
& $\mathbf{1.023 \pm 0.193}$ \\
&
&
& FDE (m)
& $\mathbf{1.066 \pm 0.184}$ \\

\midrule

\multirow[c]{4}{*}{\textit{Forecaster}}
& \multirow[c]{4}{=}{BEV GT}
& \multirow[c]{2}{=}{From-scratch}
& ADE (m)
& $\mathbf{0.469 \pm 0.197}$ \\
&
&
& FDE (m)
& $\mathbf{0.808 \pm 0.312}$ \\

\cmidrule(lr){3-5}

&
& \multirow[c]{2}{=}{FT - Tianjin sensing setup}
& ADE (m)
& $0.485 \pm 0.224$ \\
&
&
& FDE (m)
& $0.836 \pm 0.355$ \\

\cmidrule(lr){3-5}

&
& \multirow[c]{2}{=}{FT - TUMTraf sensing setup}
& ADE (m)
& $0.474 \pm 0.202$ \\
&
&
& FDE (m)
& $0.812 \pm 0.319$ \\

\cmidrule(lr){2-5}

\multirow[c]{6}{*}{\textit{Forecaster}}
& \multirow[c]{6}{=}{\textit{Projection} output from GT association}
& \multirow[c]{2}{=}{From-scratch}
& ADE (m)
& $2.647 \pm 0.600$ \\
&
&
& FDE (m)
& $3.219 \pm 0.704$ \\

\cmidrule(lr){3-5}

&
& \multirow[c]{2}{=}{FT - Tianjin sensing setup}
& ADE (m)
& $2.091 \pm 0.355$ \\
&
&
& FDE (m)
& $2.516 \pm 0.469$ \\

\cmidrule(lr){3-5}

&
& \multirow[c]{2}{=}{FT - TUMTraf sensing setup}
& ADE (m)
& $\mathbf{1.450 \pm 0.376}$ \\
&
&
& FDE (m)
& $\mathbf{1.903 \pm 0.426}$ \\

\cmidrule(lr){2-5}

\multirow[c]{6}{*}{\textit{Forecaster}}
& \multirow[c]{6}{=}{\textit{Projection} output from \textit{Matcher}}
& \multirow[c]{2}{=}{From-scratch}
& ADE (m)
& $2.660 \pm 0.585$ \\
&
&
& FDE (m)
& $3.238 \pm 0.686$ \\

\cmidrule(lr){3-5}

&
& \multirow[c]{2}{=}{FT - Tianjin sensing setup}
& ADE (m)
& $2.076 \pm 0.374$ \\
&
&
& FDE (m)
& $2.496 \pm 0.498$ \\

\cmidrule(lr){3-5}

&
& \multirow[c]{2}{=}{FT - TUMTraf sensing setup}
& ADE (m)
& $\mathbf{1.475 \pm 0.352}$ \\
&
&
& FDE (m)
& $\mathbf{1.910 \pm 0.406}$ \\

\bottomrule
\end{tabularx}
\end{table}

\begin{figure}[h]
\centering
\includegraphics[width=0.6\textwidth]{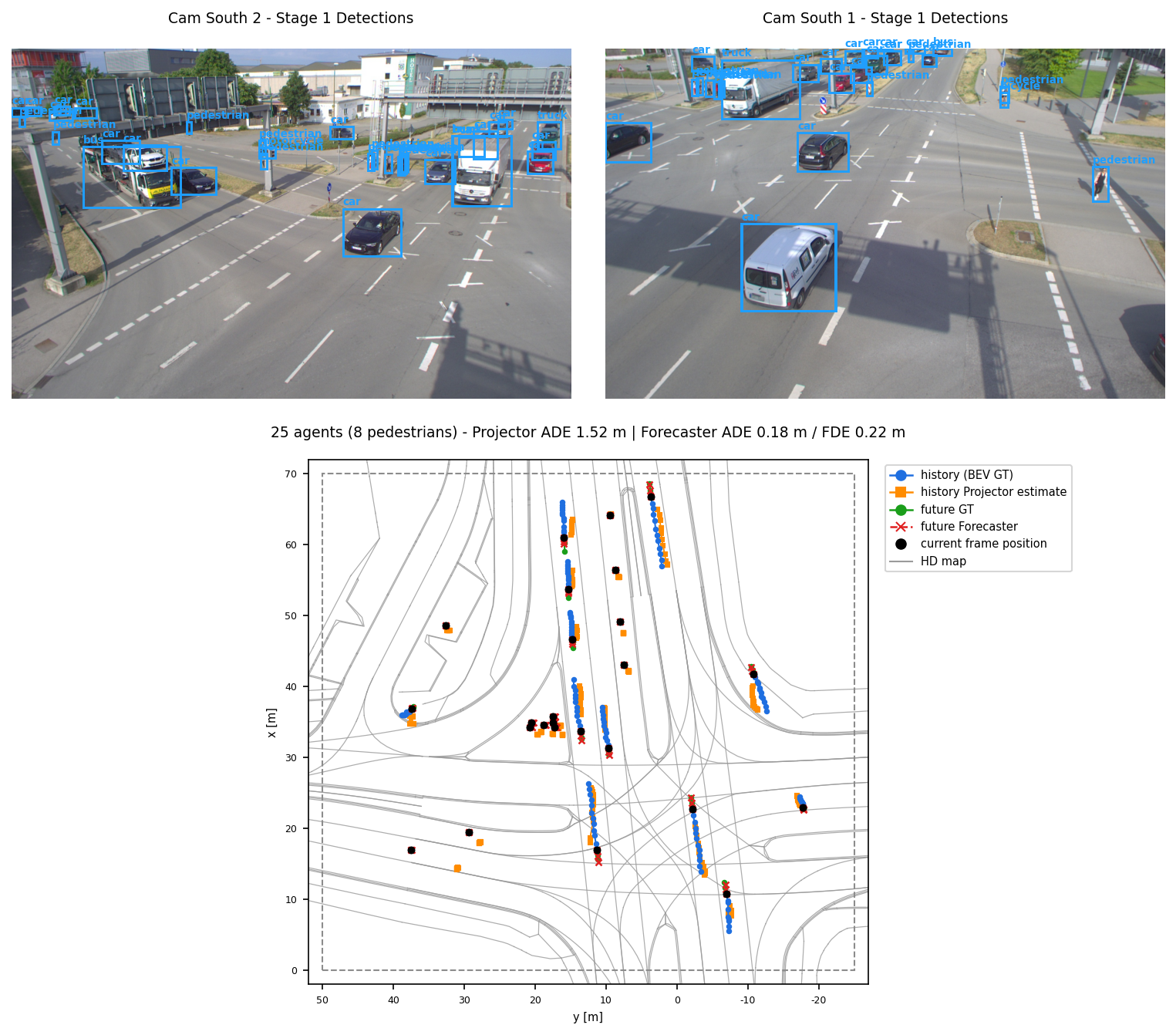}
\caption{Qualitative TUMTraf-V2X examples showing observed BEV histories, projected histories, and 1.2~s forecasts under the cascaded configuration.}
\label{fig:risultati_tumtraf}
\end{figure}

\subsection{External Roadside-Domain Evaluation on TUMTraf-V2X}

Table~\ref{tab:tumtraf_detection} reports the \textit{detector} performance (0.671 F1 overall), which should be interpreted as a lower-bound estimate because the evaluation relies on LiDAR-derived references that do not always reflect per-view visibility. As shown in Fig.~\ref{fig:stage1_gt_comparison_tumtraf}, objects included in the 3D reference may be occluded or weakly visible from a single camera (as explained in Section~\ref{sec:dataset_tumtraf}), while projected cuboids may differ from the actual image silhouette. Vulnerable road users remain the most challenging classes because of their smaller apparent size and partial visibility.

Table~\ref{tab:tumtraf_stage_results} summarizes the results by stage, input source, and training mode. The \textit{matcher} preserves strong consistency in the real domain, reaching 0.991 F1 with ground-truth projected cuboids and 0.936 with \textit{detector} outputs, confirming that cross-view association is not the limiting component. 

The \textit{projector} benefits substantially from both pretraining and source-target sensing alignment. In the from-scratch setting, the network is initialized randomly and trained exclusively on TUMTraf-V2X, yielding an ADE/FDE of 2.063/2.218~m with ground-truth associations. In the first fine-tuning setting, the model is initialized from the best \textit{projector} checkpoint obtained in the experiment described in Section~\ref{subsec:exp_r3st}, reducing ADE/FDE to 1.672/1.710~m. In the geometry-matched setting, the \textit{projector} is instead initialized from a R3ST++ checkpoint with Tianjin trajectories and intersection topology while reproducing the sensing configuration of the TUMTraf-V2X \textit{south1} and \textit{south2} cameras. This last configuration further decreases ADE/FDE to 0.978/1.038~m, showing that matching the source and target camera geometries provides an additional advantage beyond conventional checkpoint initialization. A similar trend is observed with \textit{matcher} outputs, for which the geometry-matched initialization reaches 1.023/1.066~m. Despite narrow common coverage, non-opposite viewpoints, residual synchronization effects, LiDAR-derived supervision, and only one hundred seconds of usable footage, these results show that reproducing the target sensing geometry in the synthetic source domain enables effective adaptation of the Transformer-based \textit{projector}.

On clean BEV histories, the \textit{forecaster} achieves comparable performance across training modes, with the lowest ADE/FDE obtained from scratch (0.469/0.808~m), indicating no clear benefit from initialization when accurate trajectories are already available. In the cascaded setting, performance improves as the quality of the projected histories increases: with ground-truth associations, ADE/FDE decrease from 2.647/3.219~m from scratch to 2.091/2.516~m with Tianjin initialization and 1.450/1.903~m with the geometry-matched \textit{projector}. Similar results are obtained when the \textit{projector} receives \textit{matcher} associations, yielding 2.660/3.238, 2.076/2.496, and 1.475/1.910~m, respectively. The small gap between the two association sources confirms that matching errors have limited impact, while the additional improvement mainly stems from the more accurate BEV histories produced by the sensing-aligned \textit{projector}. A qualitative example of the results obtained is shown in Fig.~\ref{fig:risultati_tumtraf}. 

Overall, the adaptation gap is mainly concentrated in the camera-to-BEV \textit{projector}, while matching remains robust and forecasting accuracy primarily depends on the quality of the reconstructed BEV histories.

\section{Conclusion}
\label{Conclusion}

This paper introduced OCTR2A, a real-time Opposite-Corner Trajectory Reconstruction and Anticipation Architecture for end-to-end perception and forecasting at signalized urban intersections. Using two synchronized roadside RGB cameras positioned at opposing corners, the proposed system recovers observed road-user motion in metric BEV coordinates and predicts its future evolution without relying on LiDAR, aerial imagery, HD maps, traffic-signal feeds, or manually calibrated homographies at inference time. Its modular design comprises four independently trainable stages optimized in cascade using the data provided by R3ST++, a synthetic benchmark and deployment-oriented simulator that replays real-world trajectories while providing synchronized multi-view imagery, cross-camera correspondences, and BEV ground truth not jointly available in existing public datasets.

The experimental analysis shows that cross-view identity recovery can be solved with purely geometric cues, without appearance-based re-identification. The learned projector reduces the mean reconstruction error by 7.8$\times$ with respect to the best calibrated geometric baseline while covering both single- and dual-view visibility regimes. The cascade training, driven mainly by the proposed soft multi-scale spatial-overlap loss, is a major contributor to OCTR2A's end-to-end accuracy, enabling it to achieve an ADE/FDE of 0.154/0.241~m at 21-28 FPS on a single consumer GPU. Evaluated in isolation on ground-truth SIND histories, the forecasting module outperforms the strongest state-of-the-art methods by 66.7\%/41.7\% in ADE/FDE while producing a single deterministic trajectory and requiring neither signal-state nor map inputs. Finally, experiments on TUMTraf-V2X indicate that reproducing the target sensing geometry in R3ST++ roughly halves the projector adaptation error compared with from-scratch training, concentrating the residual domain gap in the camera-to-BEV projection stage and supporting the use of R3ST++ for site-specific deployment.

The evaluation assumes two synchronized and partially overlapping views over a locally planar intersection, a configuration that nevertheless represents a common practical deployment scenario. In addition, the adopted detection backbone currently dominates the end-to-end latency, as efficiency optimization was beyond the initial scope of the research. Future work will address these aspects by collecting a real opposite-corner dual-camera benchmark, replacing the backbone with a substantially lighter detector to enable deployment on edge devices, extending the forecaster with multimodal decoding and longer horizons, equipping R3ST++ with additional sensor modalities to make it an effective simulation benchmark for cooperative sensing architectures as well, and investigating self-calibrating projector variants that infer the sensing geometry directly from the observed streams. R3ST++ and its generation toolkit will be publicly released to support reproducibility and future research on roadside perception-to-forecasting.

\section*{Acknowledgments}
This study has been partially supported by the Italian Ministry of Enterprises and Made in Italy (Ministero delle Imprese e del Made in Italy -- MIMIT) with the project PMDI under the agreements for innovation in the automotive sector D.M. 31/12/2021 and DD 10/10/2022.
This research was partially carried out by Claudia Melis Tonti within the framework of the National Ph.D. Program funded by the Italian PNRR (Mission 4, Component 1), under Ministerial Decree No. 118/2023, in the Department of Computer, Control and Management Engineering ``Antonio Ruberti'', Sapienza University of Rome.

\bibliographystyle{unsrtnat}
\bibliography{references}

\end{document}